%% file: main.tex
\documentclass[11pt,a4paper,dvipsnames]{article}
\usepackage[hyperref]{acl2021}

\usepackage[T1]{fontenc}
\usepackage[utf8]{inputenc}

\usepackage{times}
\usepackage{latexsym}

\newcommand{\ours}{TISA}

\usepackage{microtype}

\usepackage[group-separator={,}]{siunitx}

\usepackage{graphicx}
\usepackage{amsmath}
\usepackage{amssymb} 

\usepackage{booktabs}
\newcommand\scalemath[2]{\scalebox{#1}{\mbox{\ensuremath{\displaystyle #2}}}}

\usepackage{caption}
\usepackage{subcaption}

\usepackage{enumitem} 
\usepackage{tabularx}
\usepackage{multirow}
\usepackage{listings}
\usepackage{dblfloatfix}
\usepackage{multicol}

\setlist{nolistsep} 
\usepackage{balance}

\usepackage{booktabs,subcaption,amsfonts,dcolumn}

\usepackage{fancyhdr}
\usepackage{pifont}
\newcommand{\cmark}{\ding{51}}
\newcommand{\xmark}{\ding{55}}

\newcommand{\real}{\mathbb{R}}
\newcommand{\abs}[1]{\left|#1\right|}

\aclfinalcopy 
\def\aclpaperid{***} 

\title{The Case for Translation-Invariant Self-Attention\\{}
in Transformer-Based Language Models}
\ifx\author{Ulme Wennberg \\
  Division of Speech, Music and Hearing \\ 
  KTH Royal Institute of Technology \\
  \texttt{ulme@kth.se} \\\And
  Gustav Eje Henter \\
  Division of Speech, Music and Hearing \\ 
  KTH Royal Institute of Technology \\
  \texttt{ghe@kth.se} \\}\fi
\author{
  Ulme Wennberg \quad Gustav Eje Henter \\
Division of Speech, Music and Hearing, KTH Royal Institute of Technology, Sweden\\
   {\{ulme, ghe\}@kth.se}
}
\date{}

\begin{document}
\maketitle
\thispagestyle{plain}

\input{00-abstract}
\input{01-introduction}
\input{02-background}
\input{03-analysis-of-learned-embeddings}
\input{04-model}
\input{05-experiments}
\input{06-conclusion}
\input{07-acknowledgements}

\bibliographystyle{acl_natbib}
\bibliography{acl2021}
\balance

\input{08-supplementary}

\end{document}

%% file: 00-abstract.tex
\begin{abstract}
Mechanisms for encoding positional information are central for transformer-based language models. In this paper, we analyze the position embeddings of existing language models, finding strong evidence of translation invariance, both for the embeddings themselves and for their effect on self-attention. The degree of translation invariance increases during training and correlates positively with model performance. Our findings lead us to propose translation-invariant self-attention (TISA), which accounts for the relative position between tokens in an interpretable fashion without needing conventional position embeddings. Our proposal has several theoretical advantages over existing position-representation approaches. Experiments show that it improves on regular ALBERT on GLUE tasks, while only adding orders of magnitude less positional parameters.
\end{abstract}

%% file: 01-introduction.tex
\section{Introduction}
\label{sec:intro}

The recent introduction of transformer-based language models by \citet{vaswani2017attention} has
set new benchmarks in language processing tasks such as machine translation \citep{lample2018phrasebased,gu2018metalearning,edunov2018understanding}, question answering \citep{yamada2020luke}, and information extraction \citep{wadden2019entity,lin-etal-2020-joint}.
However, because of the non-sequential and position-independent nature of the internal components of transformers,
additional mechanisms are needed to enable models to take word order into account.

\citet{liu2020learning} identified three important criteria for ideal position encoding:
Approaches should be \emph{inductive}, meaning that they can handle sequences and linguistic dependencies of arbitrary length, \emph{data-driven}, meaning that positional dependencies are learned from data, and \emph{efficient} in terms of the number of trainable parameters.
Separately, \citet{shaw2018selfattention} argued for \emph{translation-invariant} positional dependencies that depend on the relative distances between words rather than their absolute positions in the current text fragment.
It is also important that approaches be \emph{parallelizable}, and ideally also \emph{interpretable}.
Unfortunately, none of the existing approaches for modeling positional dependencies satisfy all these criteria, as shown in Table \ref{tab:tab1} and in Sec.\ \ref{sec:background}. This is true even for recent years' state-of-the-art models such as BERT \cite{devlin2019bert}, RoBERTa \cite{liu2019roberta}, ALBERT \cite{lan2020albert}, and ELECTRA \cite{clark2020electric}, which require many positional parameters but still cannot handle arbitrary-length sequences.

\input{tables/table1.tex}

This paper makes two main contributions:
First, in Sec.\ \ref{sec:analysis}, we analyze the learned position embeddings in major transformer-based language models.
Second, in Sec.\ \ref{sec:model}, we leverage our findings to propose a new positional-dependence mechanism that satisfies all desiderata enumerated above. 
Experiments verify that this mechanism can be used alongside conventional position embeddings to improve downstream performance.
%
%
Our \href{https://github.com/ulmewennberg/tisa}{code is available}.

%% file: tables/table1.tex
\newcommand{\yes}{\textcolor{ForestGreen}{\cmark}}
\newcommand{\yesss}{\textcolor{ForestGreen}{\cmark!}}
\newcommand{\no}{\textcolor{red}{\xmark}}
\newcommand{\kinda}{\textcolor{Black}{$\mathbf{\sim}$}}

\begin{table*}[!ht]
  \centering
  \footnotesize
  \begin{tabular}{@{}l|cccccc@{}}
    \toprule
    & Induct- & Data- & Parameter & Translation & Parallel & Interpret- \\
    Method & ive? & driven? & efficient? & invariant? & -izable? & able? \\
    \midrule
    Sinusoidal position embedding \cite{vaswani2017attention} & \yes & \no & \yes & \no & \yes & \no \\
    Absolute position embedding \cite{devlin2019bert} & \no & \yes & \no & \no & \yes & \no \\
    Relative position embedding \cite{shaw2018selfattention} & \no & \yes & \yes & \yes & \no & \no \\
    T5 \cite{raffel2020exploring} & \no & \yes & \yes & \yes & \yes & \yes \\
    Flow-based \cite{liu2020learning} & \yes & \yes & \yes & \no & \no & \no \\
    Synthesizer \cite{tay2020synthesizer} & \no & \yes & \yes & \no & \yes & \no \\
    Untied positional scoring \cite{ke2020rethinking} & \no & \yes & \no & \no & \yes & \no \\
    Rotary position embedding \cite{lu-2021-roformer} & \yes & \no & \yes & \yes & \yes & \no \\
    \midrule
    Translation-invariant self-attention (proposed) & \yes & \yes & \yes & \yes & \yes & \yes \\
    \bottomrule
  \end{tabular}
  \caption{Characteristics of position-representation approaches for different language-modeling architectures.
  }
  \label{tab:tab1}
  \vspace{-1em}
\end{table*}

%% file: 02-background.tex
\section{Background}
\label{sec:background}

Transformer-based language models \cite{vaswani2017attention}
have significantly improved modeling accuracy over previous state-of-the-art models like ELMo \cite{peters2018elmo}.
However, the non-sequential nature of transformers created a need for other mechanisms to inject positional information into the architecture. This is now an area of active research, which the rest of this section will review.

The original paper by \citet{vaswani2017attention} proposed summing each token embedding with a position embedding, and then used the resulting embedding as the input into the first layer of the model.
%
%
BERT \cite{devlin2019bert} reached improved performance training data-driven $d$-dimensional embeddings for each position in text snippets of at most $n$ tokens.
A family of models have tweaked the BERT recipe to improve performance, including RoBERTa \cite{liu2019roberta} and ALBERT \cite{lan2020albert}, where the latter has layers share the same parameters to achieve a more compact model.

All these recent data-driven approaches are restricted to fixed max sequence lengths of $n$ tokens or less (typically $n=512$).
Longformer \cite{beltagy2020longformer} showed modeling improvements by increasing $n$ to 4096, suggesting that the cap on sequence length limits performance. However, the Longformer approach also increased the number of positional parameters 8-fold, as the number of parameters scales linearly with $n$; cf.\ Table \ref{tab:numb_pos_params}.

\citet{clark2019what} and \citet{htut2019attention} analyzed BERT attention, finding some attention heads to be strongly biased to local context, such as the previous or the next token.
\citet{wang2020position} found that even simple concepts such as word-order and relative distance can be hard to extract from absolute position embeddings.
\citet{shaw2018selfattention} independently proposed using relative position embeddings that depend on the signed distance between words instead of their absolute position, making local attention easier to learn.
They reached improved BLEU scores in machine translation, but their approach (and refinements by \citet{huang-2019-music-transformer}) are hard to parallelize,
which is unattractive in a world driven by parallel computing.
\citet{zeng2020prosody} used relative attention in speech synthesis, letting each query interact with separate matrix transformations for each key vector, depending on their relative-distance offset.
\citet{raffel2020exploring}
directly model
position-to-position interactions, by splitting relative-distance offsets into $q$ bins%
.
These relative-attention approaches all facilitate processing sequences of arbitrary length, but can only resolve linguistic dependencies up to a fixed predefined maximum distance%
.

\citet{tay2020synthesizer} directly predicted both word and position contributions to the attention matrix without depending on token-to-token interactions. However,
the approach is not inductive,
as the size of the attention matrix is a fixed hyperparameter.

%

\citet{liu2020learning} used sinusoidal functions with learnable parameters as position embeddings. They obtain compact yet flexible models, but use a neural ODE, which is computationally unappealing.

\citet{ke2020rethinking} showed that self-attention works better if word and position embeddings are untied to reside in separate vector spaces, but their proposal is neither inductive nor parameter-efficient.

\ifx Furthermore, \cite{liu2020learning} find that parameter sharing between different layers lead to superior results. This is something that previous models can't do because of the sheer number of trainable parameters that would be introduced. \fi

\input{figures/multifig1.tex}


%
%
\citet{lu-2021-roformer} propose rotating each embedding in the self-attention mechanism based on its absolute position, thereby inducing translational invariance, as the inner product of two vectors is conserved under rotations of the coordinate system. These rotations are, however, not learned.

The different position-representation approaches are summarized in Table \ref{tab:tab1}.
None of them satisfy all design criteria.
In this article, we analyze the position embeddings in transformer models, leading us to propose a new positional-scoring mechanism 
that combines all desirable properties (final row).

%% file: figures/multifig1.tex
\begin{figure*}
	\centering
	\begin{subfigure}[t]{0.245\textwidth}
		\centering
		\includegraphics[width=\textwidth]{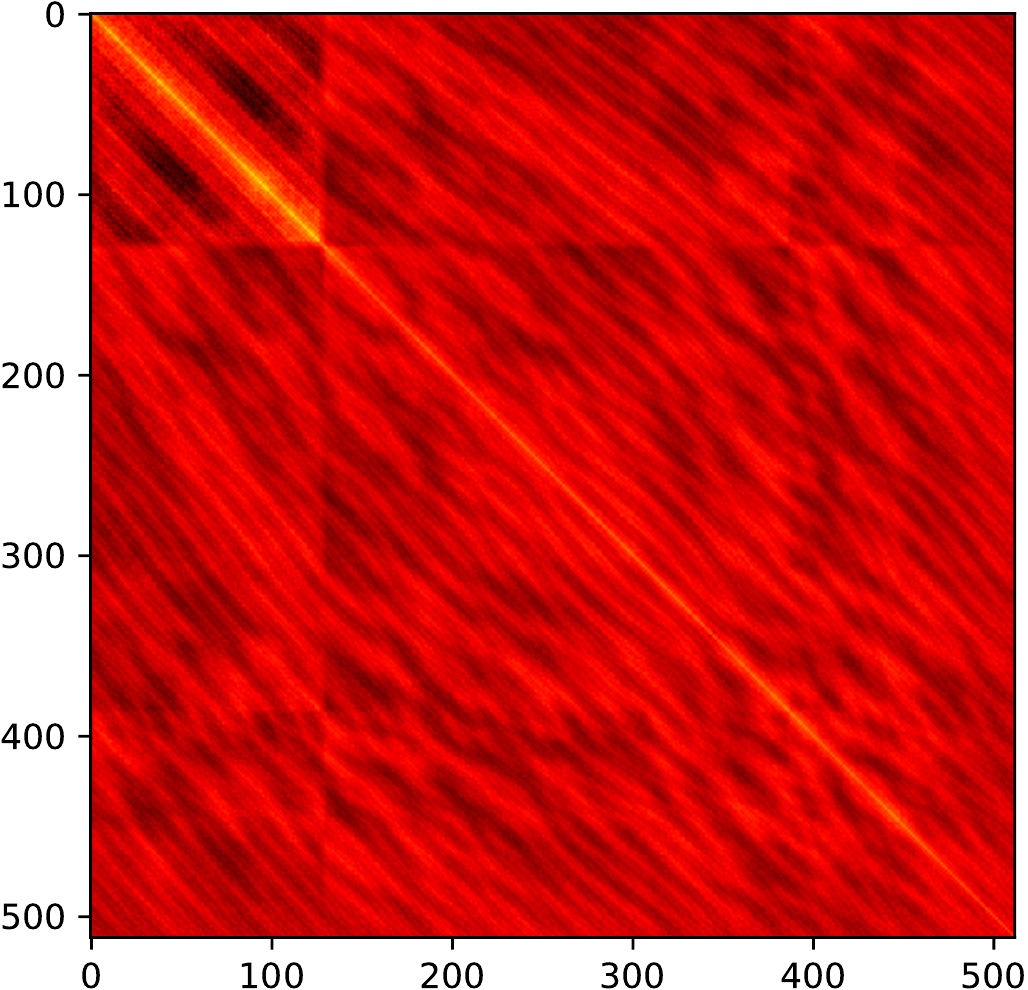}
		\caption{BERT base}
		\label{sfig:bert}
	\end{subfigure}
	\begin{subfigure}[t]{0.245\textwidth}
		\centering
		\includegraphics[width=\textwidth]{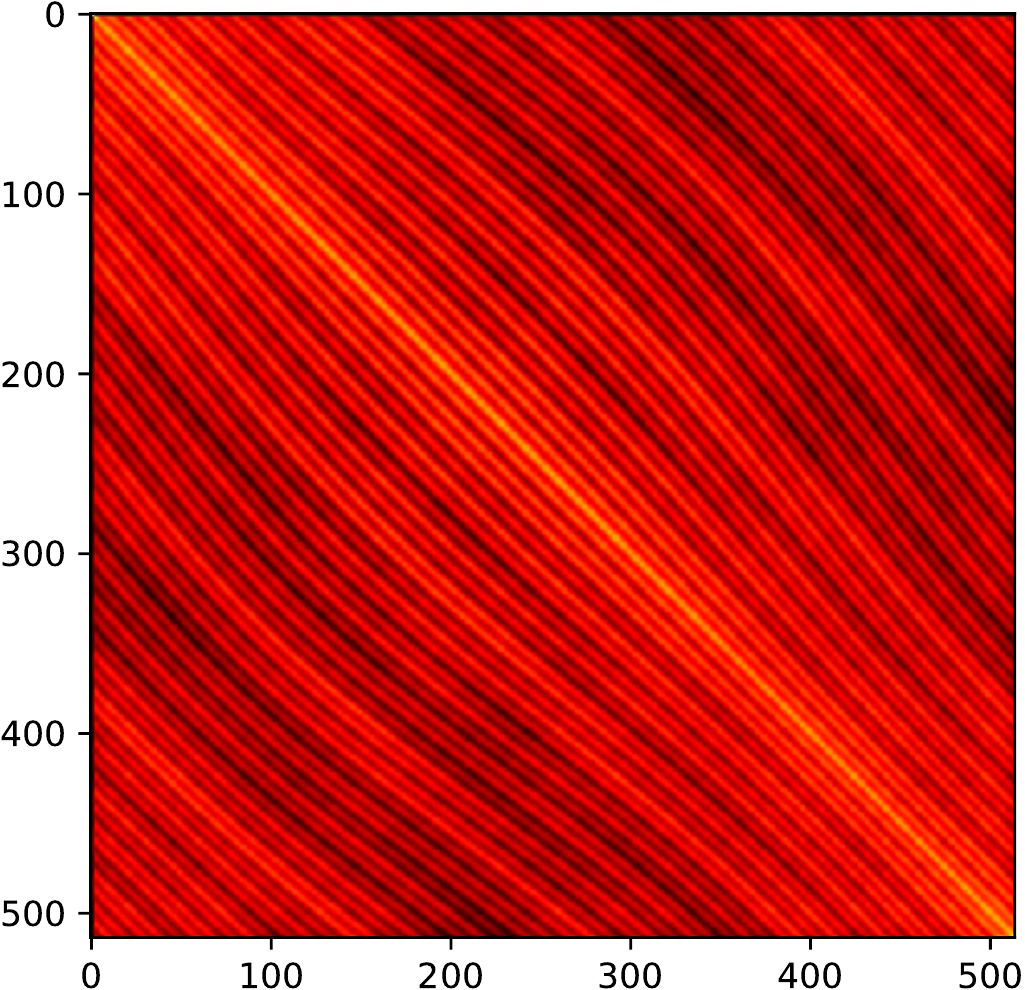}
		\caption{RoBERTa base}
		\label{sfig:roberta}
	\end{subfigure}
	\begin{subfigure}[t]{0.245\textwidth}
		\centering
		\includegraphics[width=\textwidth]{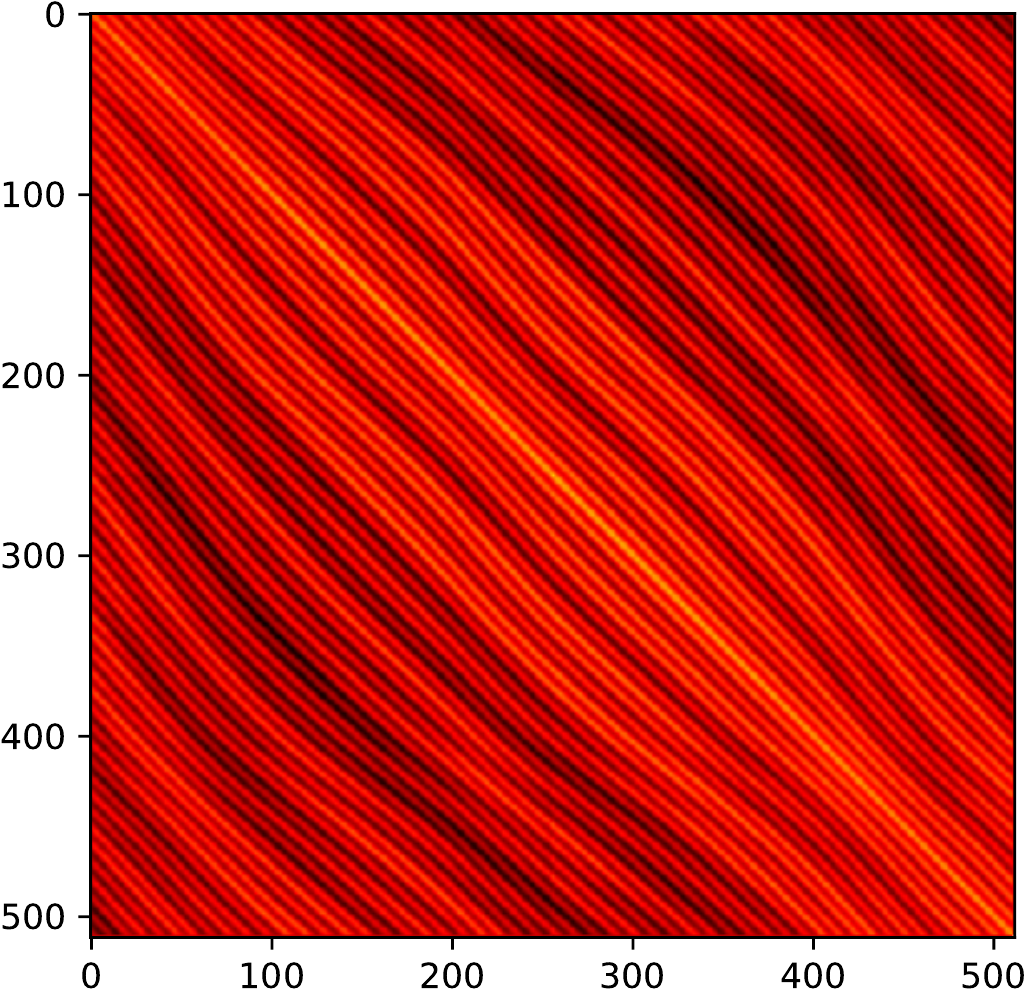}
		\caption{ALBERT base v1}
		\label{sfig:albertb}
	\end{subfigure}
	\begin{subfigure}[t]{0.245\textwidth}
		\centering
		\includegraphics[width=\textwidth]{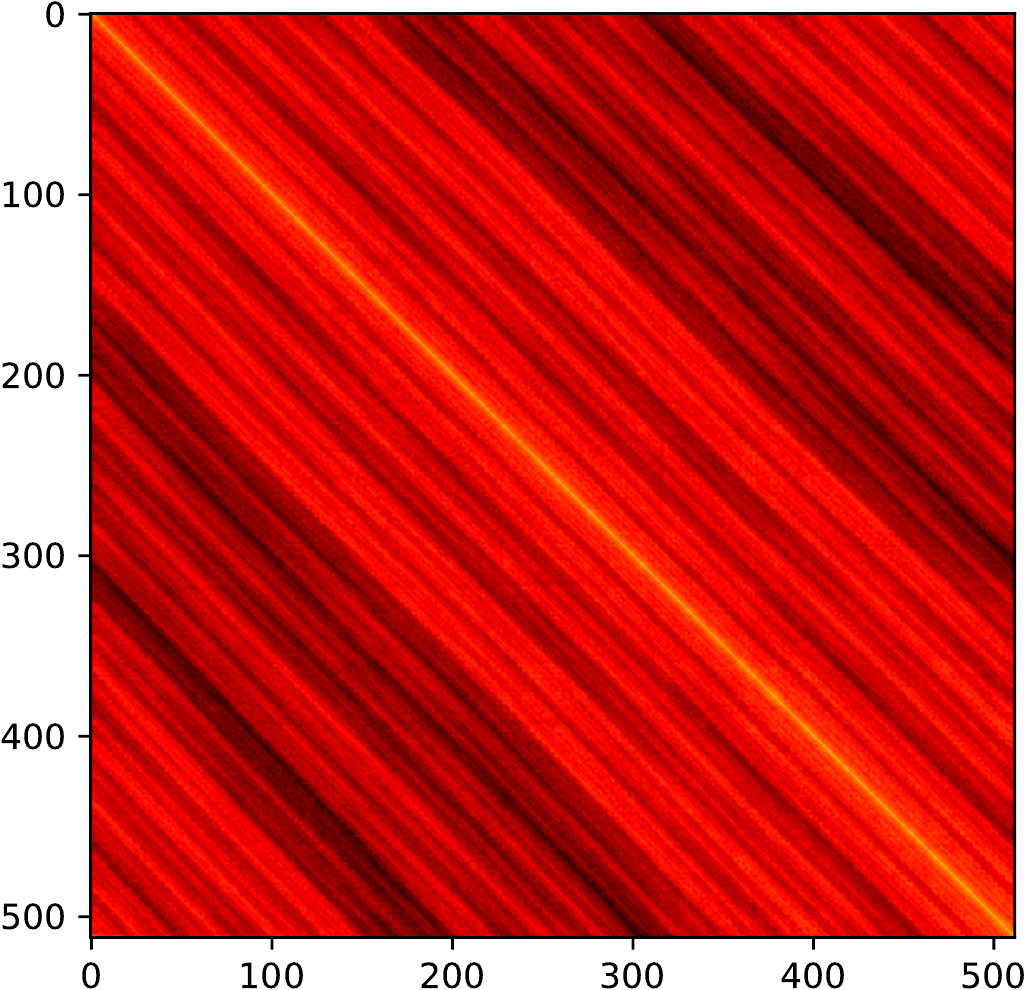}
		\caption{ALBERT xxlarge v2}
		\label{sfig:albertx}
	\end{subfigure}
	\caption{Heatmaps visualizing the matrix $P = E_P E_P^T$ of position-embedding inner products for different models. The greater the inner product between the embeddings, the brighter the color. See appendix Figs.\ \ref{fig:berts}, \ref{fig:alberts} for more.}
    \vspace{-1.5em}
	\label{fig:matrices}
\end{figure*}

%% file: 03-analysis-of-learned-embeddings.tex
\section{Analysis of Existing Language Models}
\label{sec:analysis}
In this section, we introspect selected high-profile language models to gain insight into how they have learned to account for the effect of position.

\subsection{Analysis of Learned Position Embeddings}
First, we stack the position embeddings in the matrix
$E_P \in \real^{n \times d}$, and inspect the symmetric matrix $P = E_P E_P^T \in \real^{n \times n}$
, where $P_{i,j}$ represents the inner product between the $i$th and $j$th embedding vectors.
%
%
%
If inner products are translation invariant,
$P_{i,j}$ will only depend on the difference between the indices, $j-i$,
giving a \emph{Toeplitz matrix}, a matrix where each diagonal is constant.
%

Fig.\ \ref{fig:matrices} visualizes the $P$-matrices 
for the position embeddings in a number of prominent transformer models, listed from oldest to newest, which also is in order of increasing performance.
We note that a clear Toeplitz structure emerges from left to right.
Translation invariance is also seen when plotting position-embedding cosine similarities, as done by \citet{wang2020position} for transformer-based language models and by \citet{dosovitskiy-2020-vision-transformer} for 2D transformers modeling image data.

In Fig.\ \ref{fig:albert_toeplitz_r2} we further study how the degree of Toeplitzness (quantified by $R^2$, the amount of the variance among matrix elements $P_{i,j}$ explained by the best-fitting Toeplitz matrix) changes for different ALBERT models. With longer training time (i.e., going from ALBERT v1 to v2), Toeplitzness increases, as the arrows show. This is associated with improved mean dev-set score.
Such evolution is also observed in \citet[Fig.\ 8]{wang2020position}.

\subsection{Translation Invariance in Self-Attention}
Next, we analyze how this translation invariance is reflected in self-attention.
Recall that \citet{vaswani2017attention} self-attention can be written as 
%
\begin{equation}
    \mathrm{att}(Q, K, V) = \mathrm{softmax}\left(\tfrac{QK^T}{\sqrt{d_k}}\right)V
    \text{,}
    \label{eq:attention}
\end{equation}
and define position embeddings $E_P$, word embeddings $E_W$, and query and key transformation weight matrices $W_Q$ and $W_K$. By taking
\begin{align}
    QK^T & = (E_W + E_P) W_Q W_K^T (E_W + E_P)^T
    \label{eq:qk}
\end{align}
and replacing each row of $E_W$ by the average word embedding across the entire vocabulary, we obtain a matrix we call $\widehat{F}_P$ that
quantifies the average effect of $E_P$ on the softmax in Eq.\ \eqref{eq:attention}.
Plots of the resulting $\widehat{F}_P$ for all 12 ALBERT-base attention heads in the first layer are in appendix Fig.\ \ref{fig:allheadmatrices}.
Importantly, these matrices also exhibit Toeplitz structure.
Fig.\ \ref{fig:ffunc_small} graphs sections through the main diagonal for selected heads, showing peaks at short relative distances, echoing \citet{clark2019what} and \citet{htut2019attention}.
In summary, we conclude that position encodings, and their effect on softmax attention, have an approximately translation-invariant structure in successful transformer-based language models.
%


\input{figures/fig_albert_r2}

%% file: figures/fig_albert_r2.tex
\begin{figure}[b!]
  \centering
  \includegraphics[width=\columnwidth]{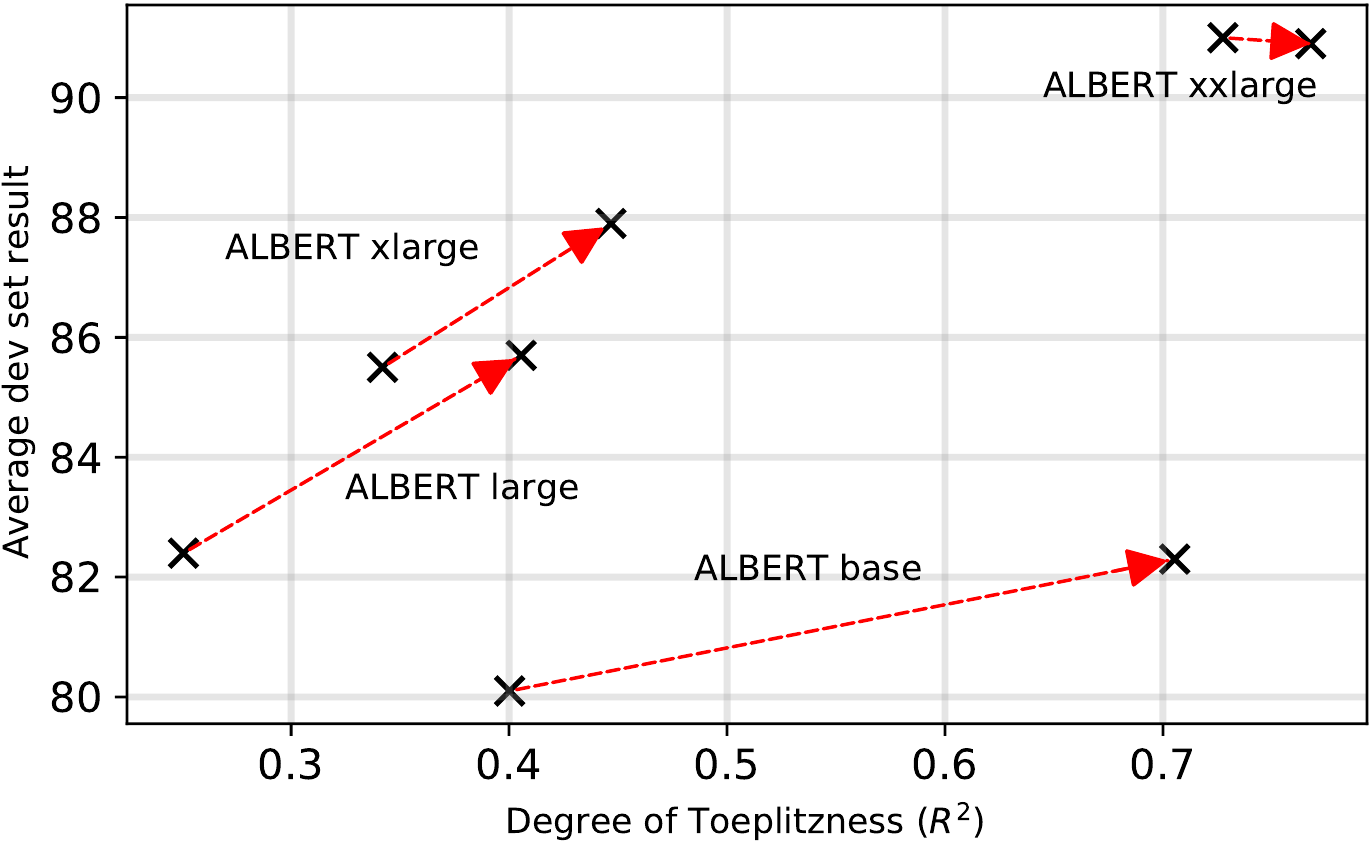}
  \vspace{-1.5em}
  \caption{Scatterplot of the degree of Toeplitzness of $P$ for different ALBERT models (v1$\rightarrow$v2) against average performance numbers (from \citeauthor{lan2020albert}'s \href{https://github.com/google-research/albert}{GitHub}) over SST-2, MNLI, RACE, and SQuAD 1.1 and 2.0.
  }
  \label{fig:albert_toeplitz_r2}
  \vspace{-1em}
\end{figure}

%% file: 04-model.tex
\section{Proposed Self-Attention Mechanism}
\label{sec:model}
We now introduce our proposal for parameterizing the positional contribution to self-attention in an efficient and translation-invariant manner, optionally removing the position embeddings entirely.



\subsection{Leveraging Translation Invariance for Improved Inductive Bias} Our starting point is the derivation of \citet{ke2020rethinking}. They expand $QK^T$ while ignoring cross terms, yielding
\begin{align}
    QK^T & \approx E_W W_Q W_K^T E_W^T + E_P W_Q W_K^T E_P^T
    \text{,}
    \label{eq:factored}
\end{align}
an approximation they support by theory and empirical evidence.
They then ``untie'' the effects of words and positions by using different $W$-matrices for the two terms in Eq.\ \eqref{eq:factored}.
We agree with separating these effects,
but also see a chance to reduce the number of parameters.

\input{figures/fig5}

Concretely, we propose to add a second term $F_P \in \real ^{n \times n}$, a Toeplitz matrix, inside the parentheses of Eq.\ \eqref{eq:attention}.
$F_P$ can either a) supplement or b) replace the effect of position embeddings on attention in our proposed model.
For case a), we simply add $F_P$ to the existing expression inside the softmax, while for case b) a term $\sqrt{d_k}F_P$ is inserted in place of the term $E_P W_Q W_K^T E_P^T$ in Eq.\ \eqref{eq:factored}. This produces two new self-attention equations:
%
\ifx
Case A:
\begin{align}
    \mathrm{softmax}\left(\tfrac{Q K^T}{\sqrt{d_k}} + F_P\right)V
    \text{,}
    \label{eq:tisa}
\end{align}
Case B:
\begin{align}
    \mathrm{softmax}\left(\tfrac{Q_W K_W^T}{\sqrt{d_k}} + F_P\right)V_W
    \text{,}
    \label{eq:tisa}
\end{align}
\fi
\ifx
\begin{multicols}{2}
  \begin{equation}
    \mathrm{softmax}\left(\tfrac{Q K^T}{\sqrt{d_k}} + F_P\right)V
  \end{equation}
  \break
  \begin{equation}
    \mathrm{softmax}\left(\tfrac{Q_W K_W^T}{\sqrt{d_k}} + F_P\right)V_W
    \text{,}
  \end{equation}
\end{multicols}
\fi
\ifx
\begin{center}
\begin{tabular}{ l l }
 \begin{equation}
    \mathrm{softmax}\left(\tfrac{Q K^T}{\sqrt{d_k}} + F_P\right)V
    \text{,}
  \end{equation} & \begin{equation}
    \mathrm{softmax}\left(\tfrac{Q_W K_W^T}{\sqrt{d_k}} + F_P\right)V_W
    \text{,}
  \end{equation} \\  
\end{tabular}
\end{center}
\fi
\begin{equation}
    \scalemath{1.0}{\mathrm{att} \! = \! \begin{cases}
                        \mathrm{softmax}\left(\tfrac{Q K^T}{\sqrt{d_k}}\! +\! F_P\right)\!V &\text{a)} \\ \mathrm{softmax}\left(\tfrac{Q_W K_W^T}{\sqrt{d_k}} \! + \! F_P\right)\!V_W &\text{b)}
                \end{cases}}
    \label{eq:tisa}
\end{equation}
%
where the inputs $Q_W$, $K_W$, and $V_W$ (defined by 
$Q_W = E_W W_Q$, and similarly for $K_W$ and $V_W$)
do not depend on the position embeddings $E_P$.
Case a) is not as interpretable as \ours{} alone (case b), since the resulting models have two terms, $E_P$ and $F_P$, that share the task of modeling positional information.
Our two proposals apply to any sequence model with a self-attention that follows Eq.\ \eqref{eq:attention}, where the criteria in Table \ref{tab:tab1} are desirable.

\subsection{Positional Scoring Function}
\ifx
\begin{equation}
    \scalemath{0.7}{F_P = \begin{bmatrix} f_0 & f_1 & \cdots & f_{n-1} \\
                          f_{-1} & f_0 & \cdots & f_{n-2} \\
                          \vdots & \vdots & \ddots & \vdots \\
                          f_{-(n-1)} & f_{-(n-2)} & \cdots & f_0
           \end{bmatrix}
           \text{.}}
    \label{eq:toeplitz}
\end{equation}
\fi

Next, we propose to parameterize the Toeplitz matrix $F_P$ using a \emph{positional scoring function} $f_{\theta}(\cdot)$ on the integers $\mathbb{Z}$, such that $(F_P)_{i,j}\!=\!f_{\theta}(j-i)$.
$f_{\theta}$ defines $F_P$-matrices of any size $n$.
The value of $f_{\theta}(j-i)$ directly models the positional contribution for how
the token at position $i$ attends to position $j$.
We call this \emph{translation-invariant self-attention}, or \ours{}.
\ours{} is inductive and
can be simplified
down to arbitrarily few trainable parameters.

Let $k=j-i$. Based on our findings for $\widehat{F}_P$ in Sec.\ \ref{sec:analysis}, we seek a parametric family $\{f_{\theta}\}$ that allows both localized and global attention, without diverging as $\abs{k} \to \infty$.
We here study one family that satisfies the criteria: %
the radial-basis functions
\begin{equation}
    f_{\theta}\left(k\right) = \sum\nolimits_{s=1}^S a_s \exp\left(- \abs{b_s} \left(k - c_s\right) ^2\right)
    \text{.}
    \label{eq:scorefunc}
\end{equation}
Their trainable parameters are $\theta=\{a_s,b_s,c_s\}_{s=1}^S$, i.e., 3 trainable parameters per kernel $s$.
Since these kernels are continuous functions (in contrast to the discrete bins of \citet{raffel2020exploring}), predictions change smoothly with distance, which seems intuitively meaningful for good generalization.

\citet{lin-etal-2019-open} found that word-order information in BERTs position embeddings gets increasingly washed out from layer 4 onward.
As suggested by \citet{dehghani2019universal} and \citet{lan2020albert}, we inject positional information into each of the $H$ heads at all $L$ layers,
resulting in one learned function $f_{\theta^{(h,l)}}$ for each head and layer.
The total number of positional parameters of \ours{} is then $3 S H L$.
As seen in Table \ref{tab:numb_pos_params}, this is several orders of magnitude less than the embeddings in prominent language models.

The inductivity and localized nature of \ours{} suggests the possibility to rapidly pre-train models on shorter text excerpts (small $n$), scaling up to longer $n$ later in training and/or at application time, similar to the two-stage training scheme used by \citet{devlin2019bert}, but without risking the undertraining artifacts visible for BERT at $n>128$ in Figs.\ \ref{fig:matrices} and \ref{fig:berts}.
However, we have not conducted any experiments on the performance of this option.


\input{tables/table2.tex}

\ifx
\subsection{Pretrain on short segments}
As noted by \cite{devlin2019bert}, attention is quadratic to the sequence length, making longer sentences disproportionately more expensive. We enhance this workflow by proposing a model that performs parameter sharing between all possible sequence lengths, making it easy to first pretrain on shorter text segments and then continue pretraining on increasingly longer text segments.

\subsection{Conditions of a satisfactory Positional Scoring Function}

Coming back to the criteria in table \ref{tab:tab1}.

\begin{enumerate}
  \item Focus on certain areas
  \item Approach constant value as the distance goes to infinity
  \item Inductive
\end{enumerate}

\emph{Highly flexible at low distances}

\emph{Approach constant value as the distance goes to infinity}
For words that are farther and farther away, the effect of each unit of additional distance should lessen the impacts of distance on each other. This would be modeled by a function that approaches a constant value as $\abs{k} \rightarrow \infty$, where $k = i-j$ is the relative distance from $w_i$ and $w_j$. The proposed function satisfies this criterion. To satisfy the flexibility criterion, the learned position scoring function should tend to a constant value for words that are far apart. Let $C$ be an arbitrary constant. Mathematically, we have that:

\begin{equation}
  \abs{k} \to \infty \implies f(k) \to C
\end{equation}

We trivially have that $f(k)$ in equation \ref{eq:scorefunc} fulfills this criterion as $\abs{f(k)} \to 0$ as $\abs{k} \to \infty$.

This means that small changes in distance at large distances will not have large impact on importance.

\emph{Inductive}
We use continuous functions to model the fact that adjacent places have similar values to introduce smoothness to our model predictions, and to enhance generalization capabilities.

\subsection{CLS-Token}

One aspect that complicates matters with using relative positional scorers is the CLS token. Normally, this token is prepended as the very first token in the sequence. However, this token will not be good to use for classification tasks as in the standard BERT model. Instead, we need to train a specific attention model over all the token representations in the final layer.

\subsection{Relative Positions give absolute information}
Since we have a CLS token

When working on this project, we sometimes heard an objection - but isn't there more information in the absolute position rather than merely the relative one?

It is true that absolute positional information provides some information to the model. However, by incorporating SEP and CLS token, the model still has the possibility to learn the absolute position when needed as it can be deduced from the relative distance to these start and stop tokens.
\fi

\ifx
Articles on why BERT has CLS token:
https://stackoverflow.com/questions/62705268/why-bert-Transformer-uses-cls-token-for-classification-instead-of-average-over

https://datascience.stackexchange.com/questions/66207/what-is-purpose-of-the-cls-token-and-why-its-encoding-output-is-important#:~:text=
\fi

%% file: figures/fig5.tex
\begin{figure}[!t]
	\centering
	\hfill
	\begin{subfigure}[t]{0.235\textwidth}
		\centering
		\includegraphics[width=\textwidth, height=1.9cm]{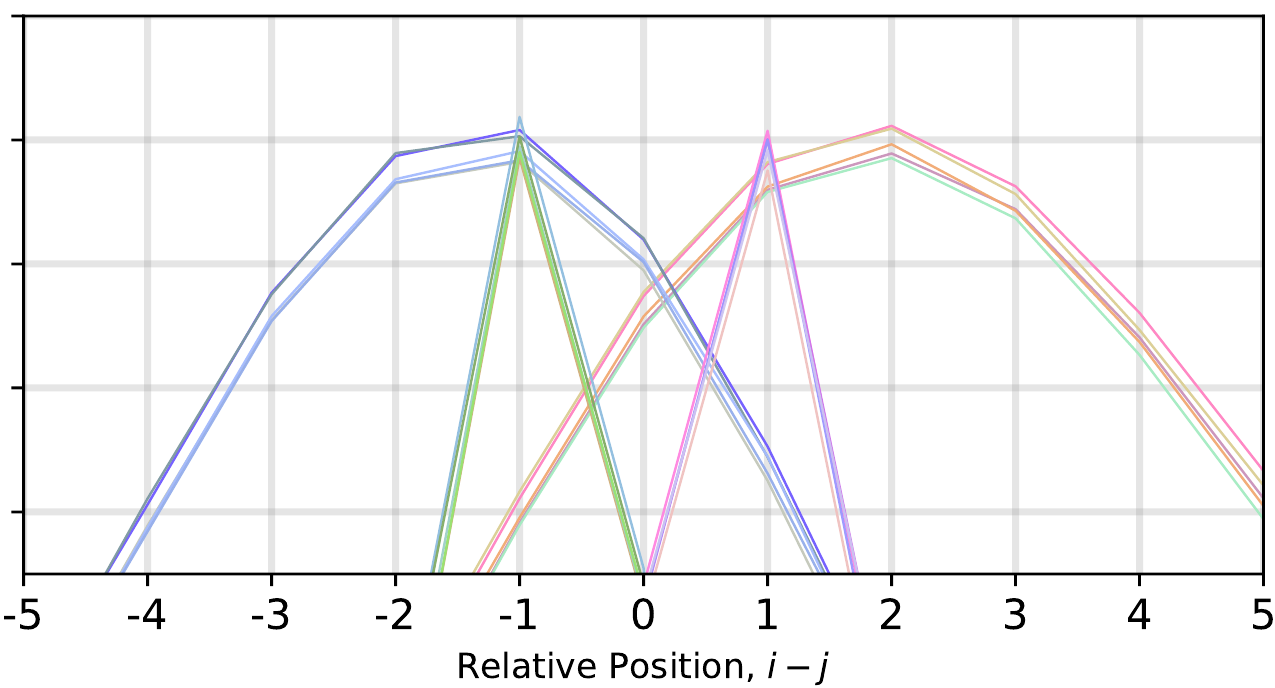}
	\end{subfigure}
	\hfill
	\hfill
	\begin{subfigure}[t]{0.235\textwidth}
		\centering
		\includegraphics[width=\textwidth, height=1.9cm]{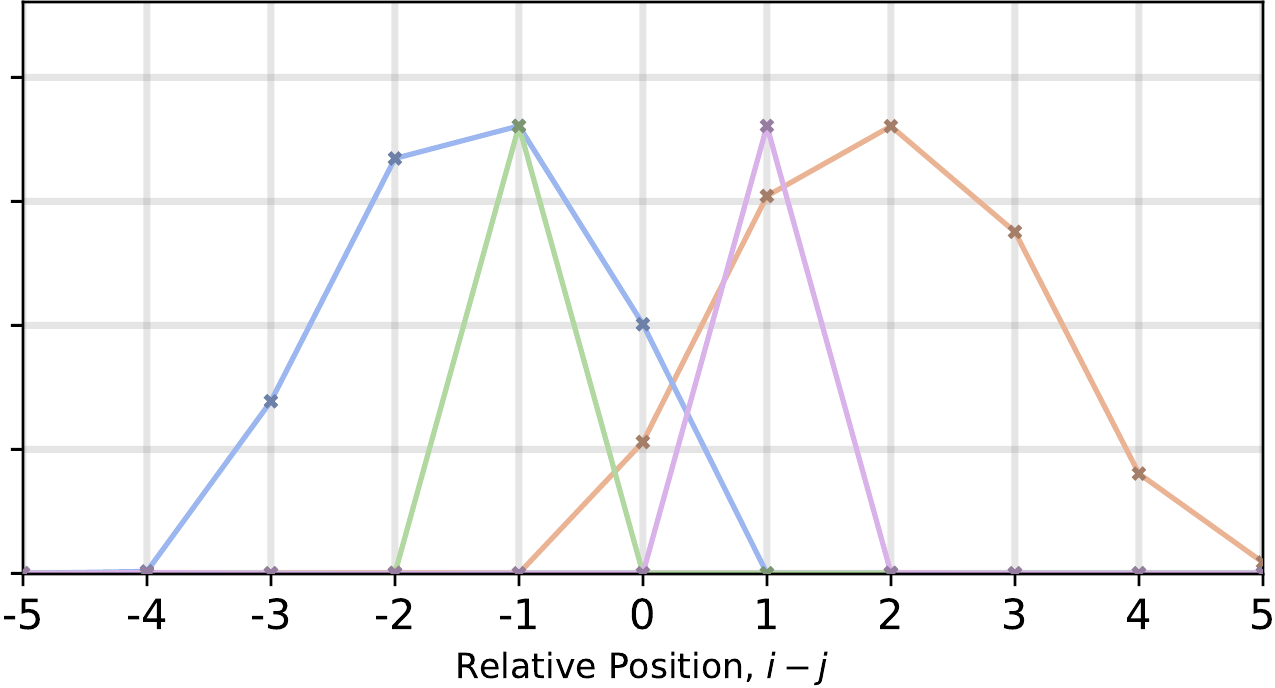}
	\end{subfigure}
	\hfill
	\caption{Positional responses of select attention heads.
	Left: Sections $(\widehat{F}_P)_{i,j}$ through $\widehat{F}_P$ of ALBERT base v2, varying $j$ for 5 different $i$, keeping $j\!=\!i$ centered. The sections are similar regardless of $i$ since $\widehat{F}_P$ is close to Toeplitz. Colors distinguish different heads.
	Right: \ours{} scoring functions, attending to similar positions as heads on the left.
	Larger plots in Figs.\ \ref{fig:albertvstisa}, \ref{fig:allheadrows}.
	}
    \label{fig:ffunc_small}
    \vspace{-1em}
\end{figure}

%% file: tables/table2.tex
\begin{table}[t]
  \centering
  \footnotesize
  \begin{tabular}{@{}l|rr|r@{}}
    \toprule
     & Standard & \citet{ke2020rethinking} & \ours{} \\
    \midrule
    General formula & $nd$ & $nd + 2d^2$ & $3SHL$ \\
    \midrule
    Longformer & \num[group-minimum-digits={3}]{3145728} & \num[group-minimum-digits={3}]{4325376} & \num[group-minimum-digits={3}, math-rm=\mathbf]{2160} \\
    BERT/RoBERTa & \num[group-minimum-digits={3}]{393216} & \num[group-minimum-digits={3}]{1572864} & \num[group-minimum-digits={3}, math-rm=\mathbf]{2 160} \\
    ALBERT & \num[group-minimum-digits={3}]{65536} & \num[group-minimum-digits={3}]{98304} & \num[group-minimum-digits={3}, math-rm=\mathbf]{2 160} \\
    \bottomrule
  \end{tabular}
  \caption{Number of positional parameters for base models
  of different language-model architectures
  and different positional information processing methods, with max sequence length $n\! \in \! (512, 4096)$, position embeddings of dimension $d \! \in \! (128, 768)$, $S\!=\!5$ kernels, $H\!=\!12$ attention heads, and $L\!=\!12$ layers with distinct \ours{} positional scoring functions.
  Parameter sharing gives ALBERT lower numbers.
  \ours{} can be used alone or added to the counts in other columns.
  }
  \label{tab:numb_pos_params}
  \vspace{-1.0em}
\end{table}

%% file: 05-experiments.tex
\section{Experiments}
\label{sec:experiments}

\input{tables/table3_and_4}
The main goal of our experiments is
to illustrate that \ours{} can be added to models to improve their performance (Table \ref{tab:results_glue_with_PE}), while adding a minuscule amount of extra parameters.
We also investigate the performance of models without position embeddings (Table \ref{tab:results_glue_without_PE}), comparing \ours{} to a bag-of-words baseline ($S=0$).
All experiments use pretrained ALBERT base v2 implemented in Huggingface \citep{wolf-etal-2020-transformers}.
Kernel parameters $\theta^{(h)}$ for the functions in Eq.\ \eqref{eq:scorefunc} were initialized by regression to the $\widehat{F}_P$ profiles of the pretrained model, (see Appendix \ref{sec:alberttotisa} for details); example plots of resulting scoring functions are provided in Fig.\ \ref{fig:ffunc_small}.
We then benchmark each configuration with and without \ours{} for 5 runs on GLUE tasks \citep{wang-etal-2018-glue}, using jiant \citep{phang2020jiant} and standard dataset splits to evaluate performance.

Our results
in Table \ref{tab:results_glue_with_PE}
show relative error reductions between 0.4 and 6.5\% when combining \ours{} and conventional position embeddings.
These gains are relatively stable regardless of $S$.
We also note that \citet{lan2020albert}\ report 92.9 on SST-2 and 84.6 on MNLI, meaning that our contribution leads to between 1.3 and 2.8\% relative error reductions over their scores.
The best performing architecture ($S\!=\!5$), gives improvements over the baseline on 7 of the 8 tasks considered and on average increases the median F1 score by 0.4 points.
All these gains have been realized using a very small number of added parameters, and without pre-training on any data after adding \ours{} to the architecture. The only joint training happens on the training data of each particular GLUE task.

Results for \ours{} alone, in Table \ref{tab:results_glue_without_PE}, are not as strong.
This could be because these models are derived from an ALBERT model pretrained using conventional position embeddings, since we did not have the computational resources to tune from-scratch pretraining of \ours{}-only language models.

Figs.\ \ref{fig:ffunc_small} and \ref{fig:albertvstisa}
plot scoring functions of different attention heads from the initialization described in Appendix \ref{sec:alberttotisa}.
Similar patterns arose consistently and rapidly in preliminary experiments on pretraining \ours{}-only models from scratch. 
The plots show heads specializing in different linguistic aspects, such as the previous or next token, or multiple tokens to either side, with other heads showing little or no positional dependence.
This
mirrors the visualizations of ALBERT base attention heads in Figs.\ \ref{fig:ffunc_small}, \ref{fig:albertvstisa}, \ref{fig:allheadrows}, \ref{fig:allheadmatrices} and
the findings of \citet{htut2019attention} and \citet{clark2019what} on BERT, but \ours{} makes this directly visible in an interpretable model, without having to probe correlations in a black box.

Interestingly, the ALBERT baseline on STS-B in Table \ref{tab:results_glue_with_PE} is only 1.3 points ahead of the bag-of-words baseline in Table \ref{tab:results_glue_without_PE}.
This agrees with experiments shuffling the order of words 
\citep{pham2020order,sinha2021masked} finding that modern language models tend to focus mainly on
higher-order word co-occurrences, rather than word order,
and suggests that word-order information is underutilized in state-of-the-art language models.

%% file: tables/table3_and_4.tex
\begin{table}
\begin{subtable}[t]{0.48\textwidth}
  \footnotesize
  \centering
\begin{tabular}{@{}l|r|rrr|r|r@{}}
    \toprule
    Task & Baseline & $S\!=\!1$ & $3$ & $5$ & $\Delta $ & $\Delta \%$ \\
    \midrule
    SST-2 & 92.9 & \textbf{93.3} & 93.1 & 93.1 & 0.4 & \hphantom{1}6.5\%\\
    MNLI & 83.8 & 84.1 & 84.4 & \textbf{84.8} & 1.0 & 5.9\%\\
    QQP & 88.2 & 88.0 & \textbf{88.3} & \textbf{88.3} & 0.1 & 1.2\%\\
    STS-B & 90.3 & \textbf{90.4} & 90.0 & \textbf{90.4} & 0.1 & 1.5\%\\
    CoLA & 57.2 & 57.0 &  56.5 & \textbf{58.5} & 1.3 & 2.9\%\\
    MRPC & 89.6 & \textbf{90.1} & 89.0 & \textbf{90.1} & 0.5 & 5.3\%\\
    QNLI & 91.6 & \textbf{91.7} & 91.4 & 91.6 & 0.1 & 0.4\%\\
    RTE & 72.9 & 71.1 & \textbf{73.6} & \textbf{73.6} & 0.7 & 2.7\%\\
    \bottomrule
  \end{tabular}
\caption{ALBERT base v2 models with position embeddings}
\vspace{1.2em}
\label{tab:results_glue_with_PE}
\end{subtable}
\hspace{\fill}
\begin{subtable}[t]{0.48\textwidth}
\flushright
\footnotesize
\centering
\begin{tabular}{@{}l|r|rrr|r|r@{}}
    \toprule
    Task & Baseline & $S\!=\!1$ & $3$ & $5$ & $ \Delta$ & $\Delta \%$ \\
    \midrule
    SST-2 & 85.1 & 85.9 & 85.8 & \textbf{86.0} & 0.9 & 6.2\%\\
    MNLI & 78.8  & 80.9 & 81.4 & \textbf{81.6} & 2.8 & 13.4\%\\
    QQP & 86.3 & 86.2 & 86.5 & \textbf{86.8} & 0.5 & 3.4\%\\
    STS-B & 89.0 & 89.0 & \textbf{89.1} & \textbf{89.1}  & 0.1 & 0.3\%\\
    MRPC & 82.8 & 83.1 & \textbf{83.3} & 83.1 & 0.5 & 3.3\%\\
    QNLI & 86.6 & 87.2 & 87.4  & \textbf{87.7} & 1.1 & 7.8\%\\
    RTE & 62.1  & 61.7 & 62.5 & \textbf{62.8} & 0.7 & 1.9\%\\
    \bottomrule
  \end{tabular}
\caption{ALBERT base v2 models without position embeddings}
\label{tab:results_glue_without_PE}
\label{tab:table1_b}
\end{subtable}
\caption{GLUE task dev-set performance (median over 5 runs) with \ours{} ($S$ kernels) and without (baseline). $\Delta$ is the maximum performance increase in a row and $\Delta \%$ is the corresponding relative error reduction rate.
  }
\vspace{-1.0em}
\end{table}

%% file: 06-conclusion.tex
\section{Conclusion}
\label{sec:conclusion}
We have analyzed state-of-the-art transformer-based language models, finding that translation-invariant behavior emerges during training. Based on this we proposed \ours{}, the first positional information processing method to simultaneously satisfy the six key design criteria in Table \ref{tab:tab1}. Experiments demonstrate competitive downstream performance.
The method is applicable also to transformer models outside language modeling, such as modeling time series in speech or motion synthesis, or to describe dependencies between pixels in computer vision.


%% file: 07-acknowledgements.tex
\section*{Acknowledgments}
We would like to thank Gabriel Skantze, Dmytro Kalpakchi, Viktor Karlsson, Filip Cornell, Oliver Åstrand, and the anonymous reviewers for their constructive feedback.
This research was partially supported by the Wallenberg AI, Autonomous Systems and Software Program (WASP) funded by the Knut and Alice Wallenberg Foundation.

%% file: 08-supplementary.tex
\newpage
\newpage
\clearpage

\appendix

\section{Visualizing $E_P E_P^T$ for Additional Language Models}
Fig.\ \ref{fig:matrices} shows the inner product between different position embeddings for the models BERT base uncased, RoBERTa base, ALBERT base v1 as well as ALBERT xxlarge v2. Leveraging our analysis findings of translation invariance in the matrix of $E_P E_P^T$ in these pretrained networks, we investigate the generality of this phenomenon by visualizing the same matrix for additional existing large language models. We find that similar Toeplitz patterns emerge for all investigated networks.

\input{figures/multifix_suppl3_EpEpT}

\section{Coefficient of Determination $R^2$}
The coefficient of determination, $R^2$, is a widely used concept in statistics that measures what fraction of the variance in a dependent variable that can be explained by an independent variable. Denoting the Residual Sum of Squares, $RSS$, and Total Sum of Squares, $TSS$, we have that
\begin{equation}
    R^2 = 1 - \tfrac{RSS}{TSS} \text{,}
\end{equation}
where $R^2\!=\!0$ means that the dependent variable is not at all explained, and $R^2\!=\!1$ means that the variance is fully explained by the independent variable. 

Applied to a matrix, $A \in \real^{n \times n}$, to determine its degree of Toeplitzness, we get $RSS$ by finding the Toeplitz matrix, $A_T \in \real^{n \times n}$, that minimizes the following expression:
\begin{equation}
    RSS = \text{min}_{A_T} \sum_{i=1}^n \sum_{j=1}^n \left( A - A_T \right)_{i,j}^2
\end{equation}
Furthermore,
we can compute $TSS$ as:
\begin{equation}
    TSS = \sum_{i=1}^n \sum_{j=1}^n \left( A_{i,j} - \left(\frac{1}{n^2} \sum_{i=1}^n \sum_{j=1}^n A_{i,j}\right) \right)^2
\end{equation}
\section{Extracting ALBERT positional scores}
\label{sec:alberttotisa}
In order to extract out the positional contributions to the attention scores from ALBERT, we disentangle the positional and word-content contributions from equation \eqref{eq:factored}, and remove any dependencies on the text sequence through $E_W$. We exchange $E_W \approx E_{\overline{W}}$, with the average word embedding over the \emph{entire vocabulary}, which we call $E_{\overline{W}}$. 
\begin{align}
    F_P & \approx \frac{1}{\sqrt{d_k}} ( E_W W_Q W_K^T E_P^T + \\
    & + E_P W_Q W_K^T E_W^T + E_P W_Q W_K^T E_P^T ) \\
    & \approx \frac{1}{\sqrt{d_k}} ( E_{\overline{W}} W_Q W_K^T E_P^T  + \\
    & + E_P W_Q W_K^T E_{\overline{W}}^T+ E_P W_Q W_K^T E_P^T )
    \label{eq:Fp_approximation}
\end{align}
This way, we can disentangle and extract the positional contributions from the ALBERT model.




\paragraph{Initialization of Position-Aware Self-Attention}
Using this trick, we initialize $F_P$ with formula \eqref{eq:Fp_approximation}. Since $F_P$ is only generating the positional scores, which are independent of context, it allows for training a separate positional scorer neural network to predict the positional contributions in the ALBERT model. Updating only 2,160 parameters (see Table \ref{tab:numb_pos_params}) significantly reduces the computational load. This pretraining initialization scheme converges in less than 20 seconds on a CPU. 



\paragraph{Removing Position Embeddings}
When removing the effect of position embeddings, we calculate the average position embedding and exchange all position embeddings for it. This reduces the variation between position embeddings, while conserving the average value of the original input vectors $E_W \!+ \! E_P$.

\paragraph{Extracted Attention Score Contributions}
Leveraging our analysis findings of translation invariance in large language models, we visualize the scoring functions as a function of relative distance offset between tokens. Fig.\ \ref{fig:ffunc_small} shows the implied scoring functions for 4 attention heads for 5 different absolute positions. Figs.\ \ref{fig:albertvstisa}, \ref{fig:allheadrows} show all 12 attention heads of ALBERT base v2 with \ours{}.

\section{Number of Positional Parameters of Language Models}
In the paper, define positional parameters as those modeling only positional dependencies. In most BERT-like models, these are the position embeddings only (typically $n \! \times \! d$ parameters). \citet{ke2020rethinking} propose to separate position and content embeddings, yielding more expressive models with separate parts of the network for processing separate information sources. In doing so, they introduce two weight matrices specific to positional information processing, $U_Q \! \in \! \real^{d \times d}$ and $U_K \! \in \! \real^{d \times d}$, totaling $nd \! +\! 2d^2$ positional parameters.

\input{figures/multifix_suppl_merge_funcs}
\input{figures/multifix_suppl5_albert_regular_att_matrix}

\paragraph{Hyperparameter Selection}
We performed a manual hyperparameter search starting from the hyperparameters that the \citet{lan2020albert} report in \url{https://github.com/google-research/albert/blob/master/run_glue.sh}. Our hyperparameter config files can be found with our code.

\section{Reproducibility}
Experiments were run on a GeForce RTX 2080 machine with 8 GPU-cores. Each downstream experiment took about 2 hours to run.
Datasets and code can be downloaded from \url{https://github.com/nyu-mll/jiant/blob/master/guides/tasks/supported_tasks.md} and \url{https://github.com/ulmewennberg/tisa}.

%% file: figures/multifix_suppl3_EpEpT.tex
\begin{figure*}[ht!]
	\centering
	\begin{subfigure}[t]{0.245\textwidth}
		\centering
		\includegraphics[width=\textwidth]{images/EpEpT/bert-base-uncased-crop.pdf}
	    \caption{BERT base uncased}
	\end{subfigure}
	\begin{subfigure}[t]{0.245\textwidth}
		\centering
		\includegraphics[width=\textwidth]{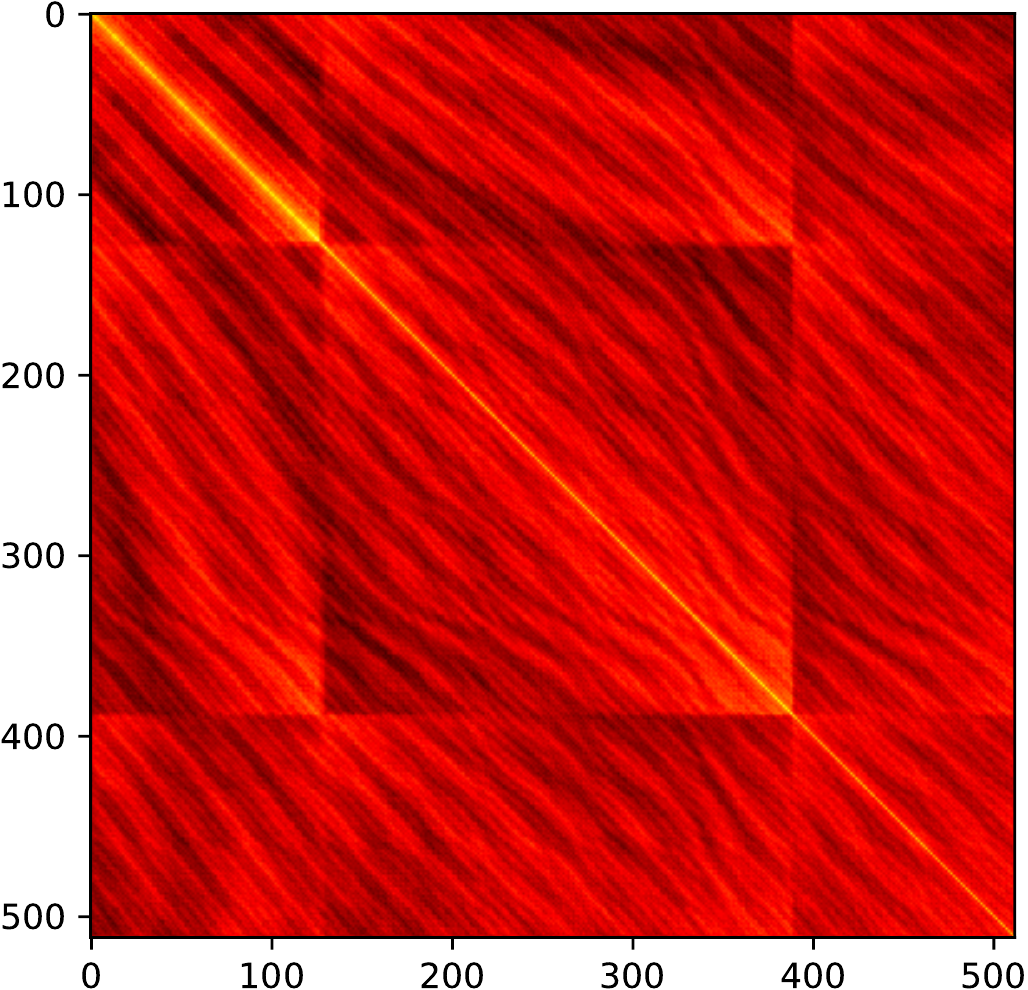}
	    \caption{BERT large uncased}
	\end{subfigure}
	\begin{subfigure}[t]{0.245\textwidth}
		\centering
		\includegraphics[width=\textwidth]{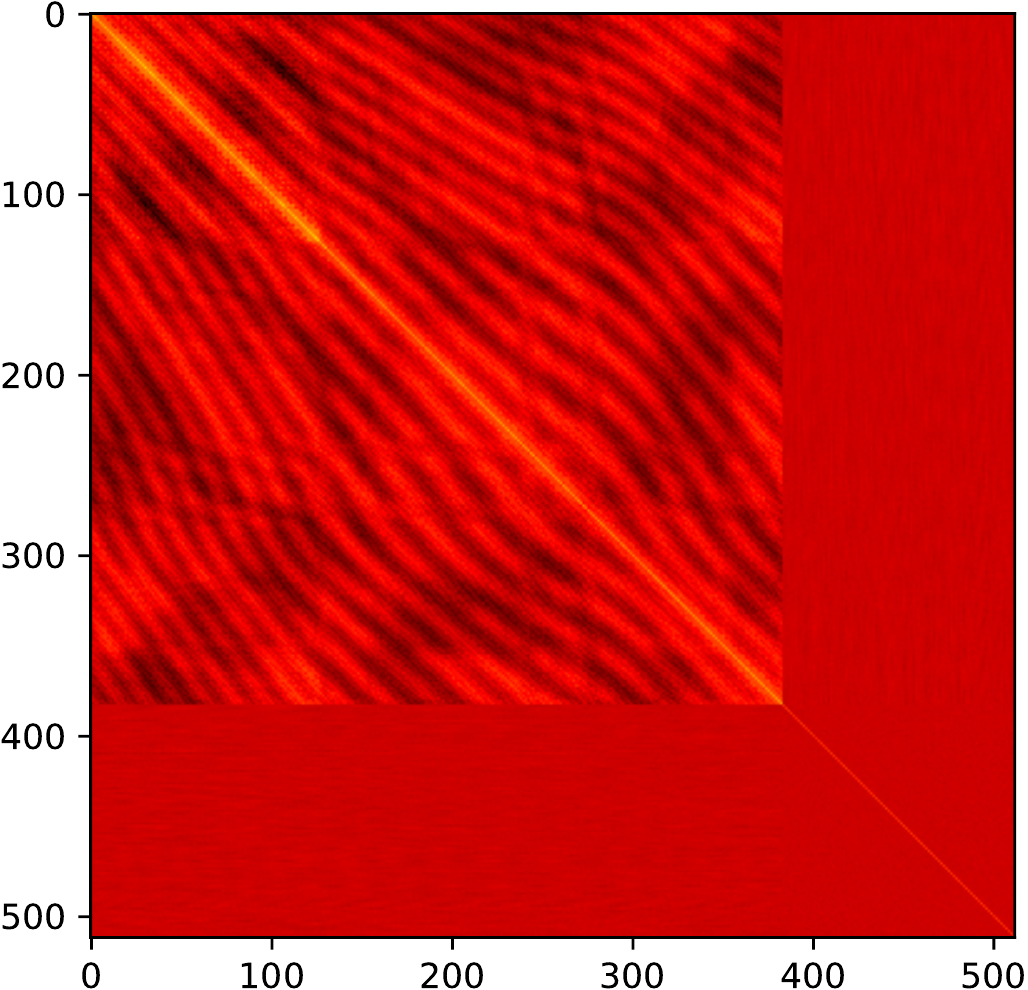}
	    \caption{BERT base cased}
	\end{subfigure}
	\begin{subfigure}[t]{0.245\textwidth}
		\centering
		\includegraphics[width=\textwidth]{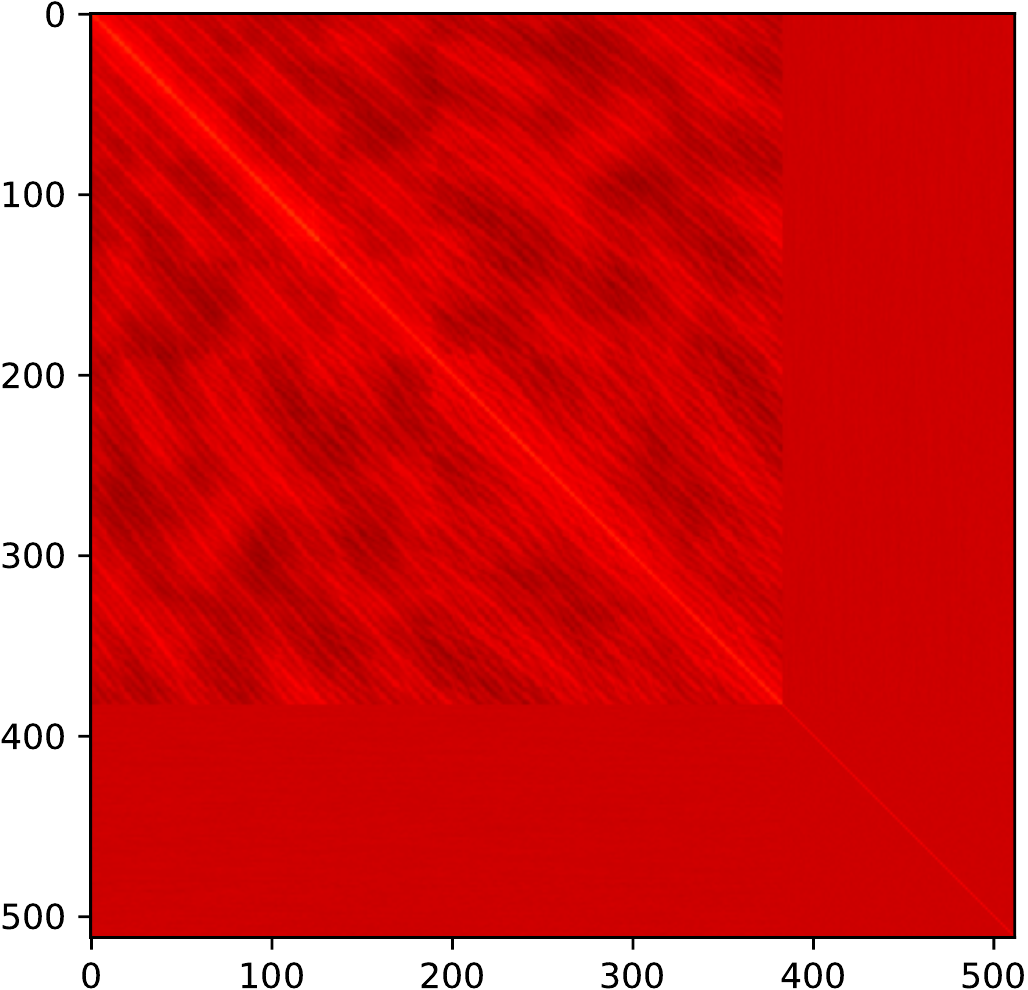}
	    \caption{BERT large cased}
	\end{subfigure}
	\begin{subfigure}[t]{0.245\textwidth}
		\centering
		\includegraphics[width=\textwidth]{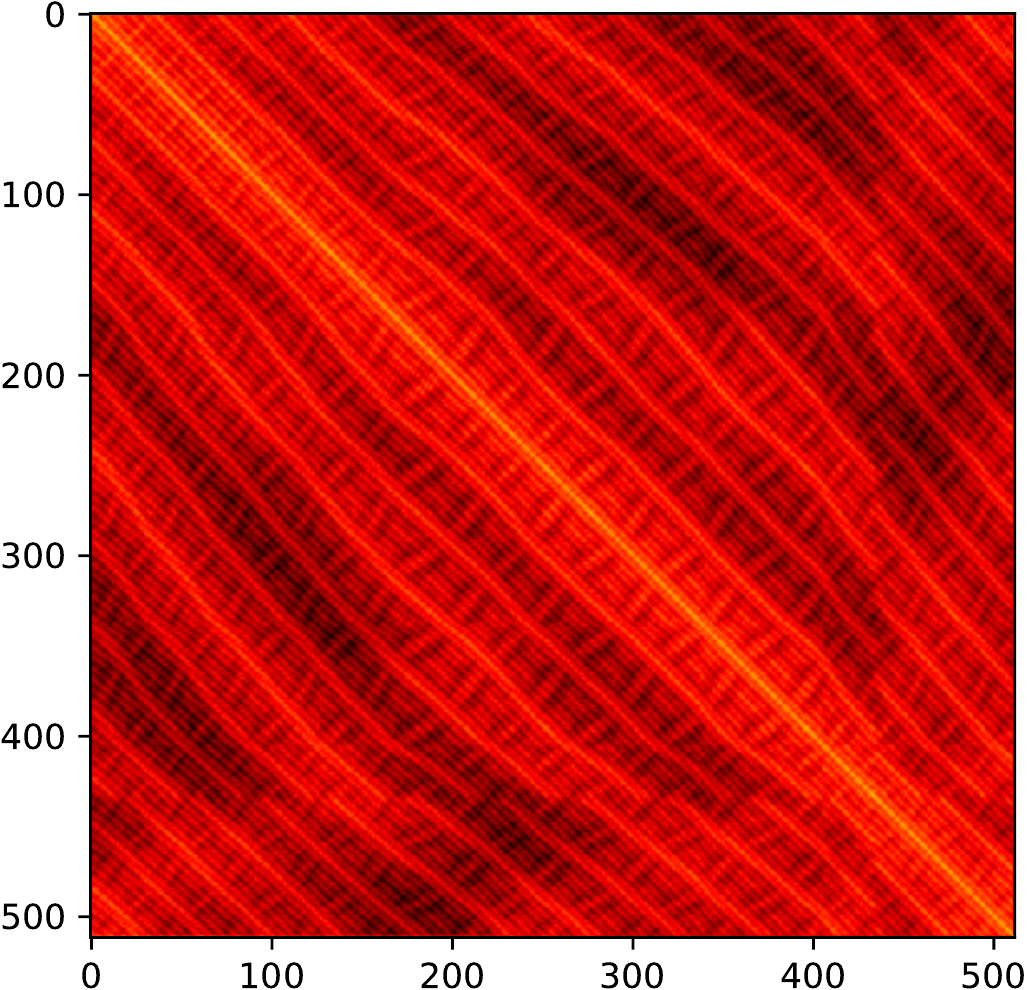}
	    \caption{ELECTRA small}
	\end{subfigure}
	\begin{subfigure}[t]{0.245\textwidth}
		\centering
		\includegraphics[width=\textwidth]{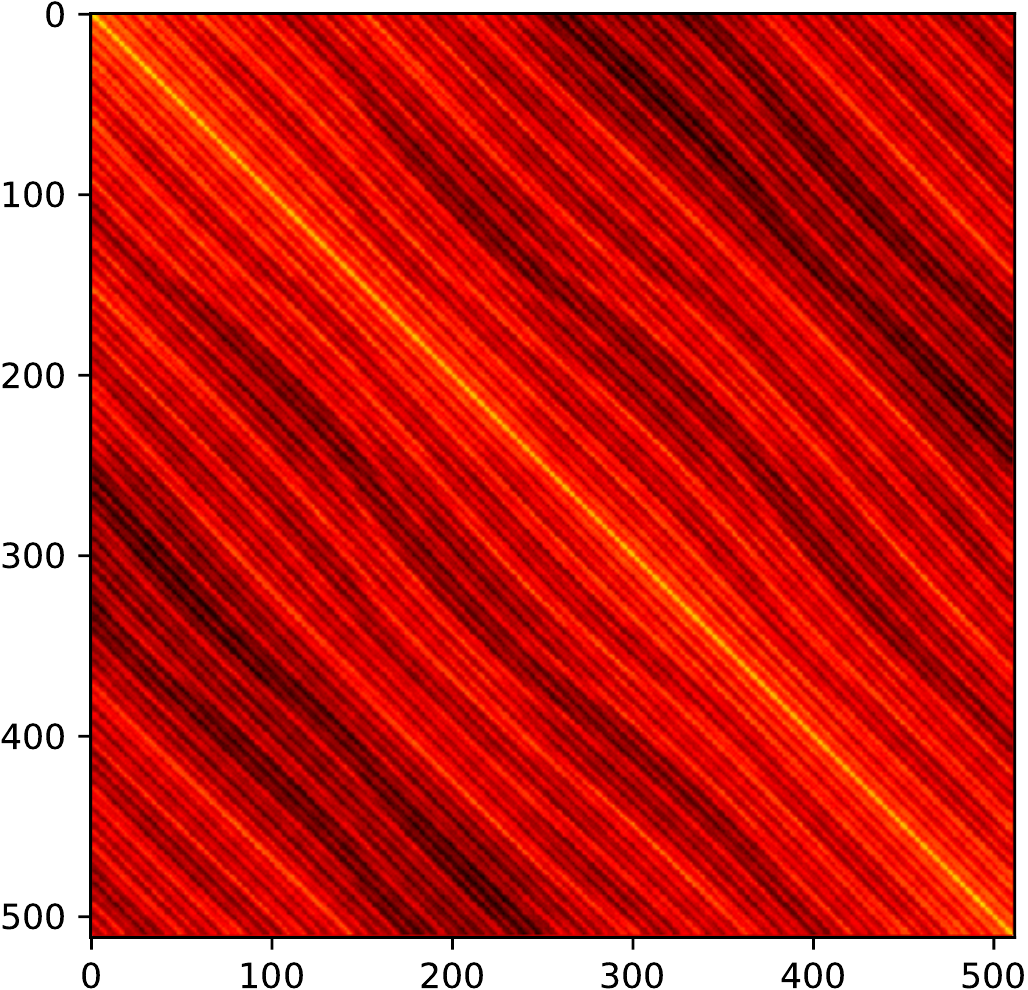}
	    \caption{ELECTRA large}
	\end{subfigure}
	\begin{subfigure}[t]{0.245\textwidth}
		\centering
		\includegraphics[width=\textwidth]{images/EpEpT/roberta-base-crop.pdf}
	    \caption{RoBERTa base}
	\end{subfigure}
	\begin{subfigure}[t]{0.245\textwidth}
		\centering
		\includegraphics[width=\textwidth]{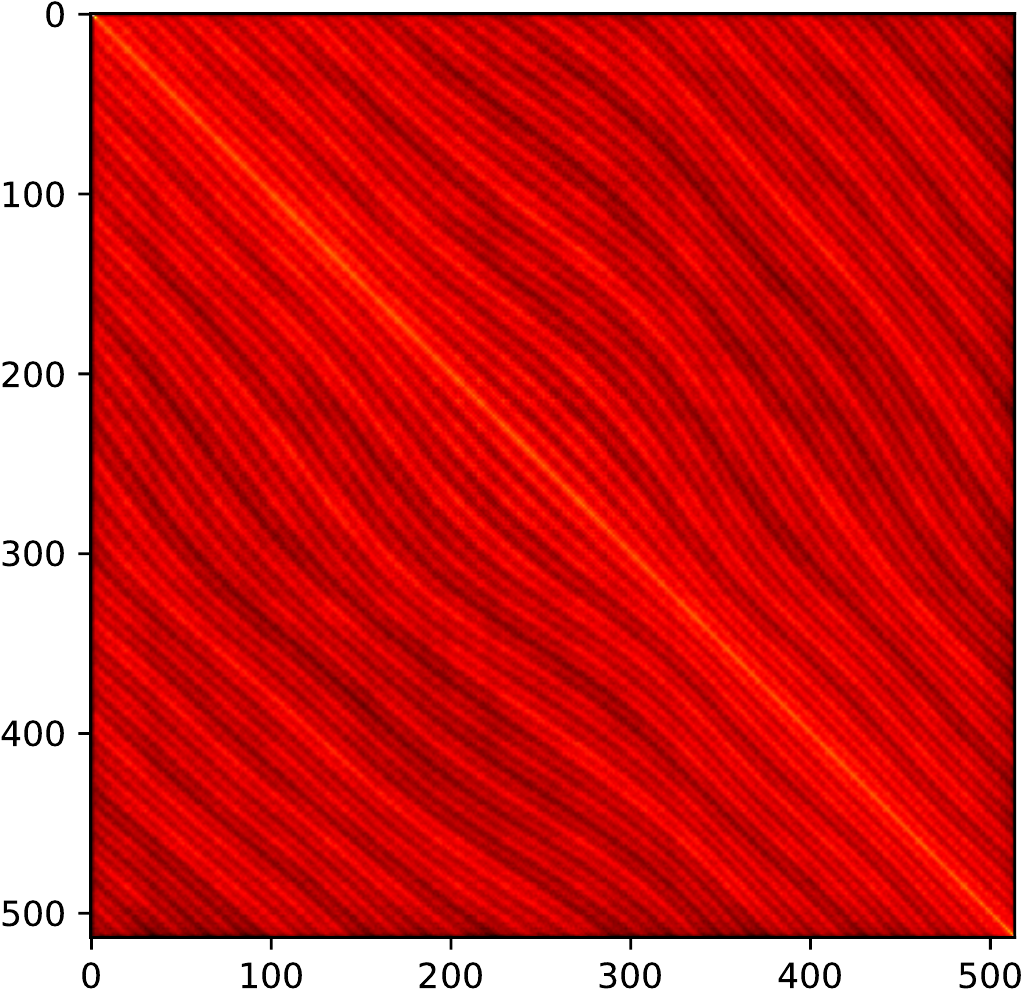}
	    \caption{RoBERTa large}
	\end{subfigure}
	\caption{Visualizations of the inner-product matrix $P = E_P E_P^T \in \real^{n \times n}$ for different BERT, ELECTRA, and RoBERTa models. We see that  ELECTRA and RoBERTa models show much stronger signs of translational invariance than their BERT counterparts. Most BERT models follow the pattern noted by \citet{wang2020position}, where the Toeplitz structure is much more pronounced for the first $128 \times 128$ submatrix, reflecting how these models mostly were trained on 128-token sequences, and only scaled up to $n=512$ for the last 10\% of training \citep{devlin2019bert}. Position embeddings 385 through 512 of the BERT cased models show a uniform color, suggesting that these embeddings are almost completely untrained.}
	\label{fig:berts}
	\centering
	\begin{subfigure}[t]{0.245\textwidth}
		\centering
		\includegraphics[width=\textwidth]{images/EpEpT/albert-base-v1-crop.pdf}
	    \caption{ALBERT base v1}
	\end{subfigure}
	\begin{subfigure}[t]{0.245\textwidth}
		\centering
		\includegraphics[width=\textwidth]{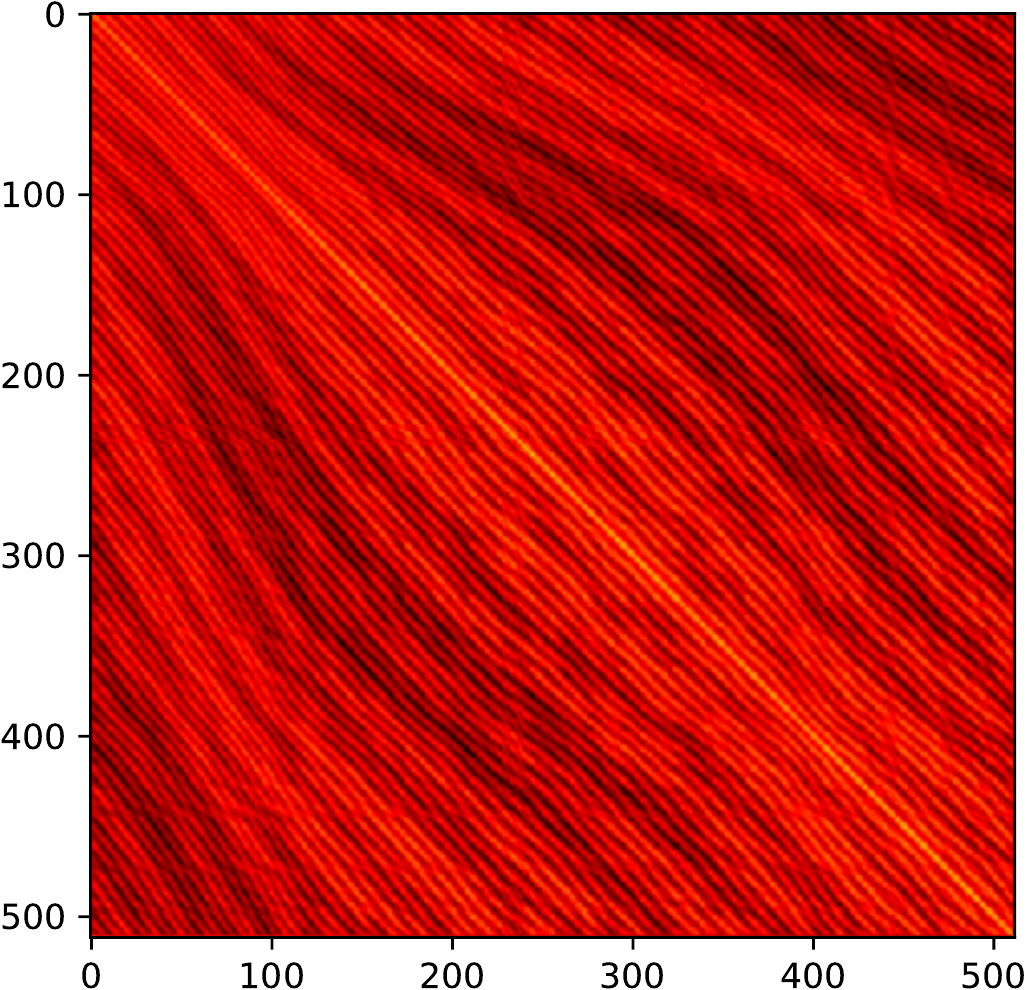}
	    \caption{ALBERT large v1}
	\end{subfigure}
	\begin{subfigure}[t]{0.245\textwidth}
		\centering
		\includegraphics[width=\textwidth]{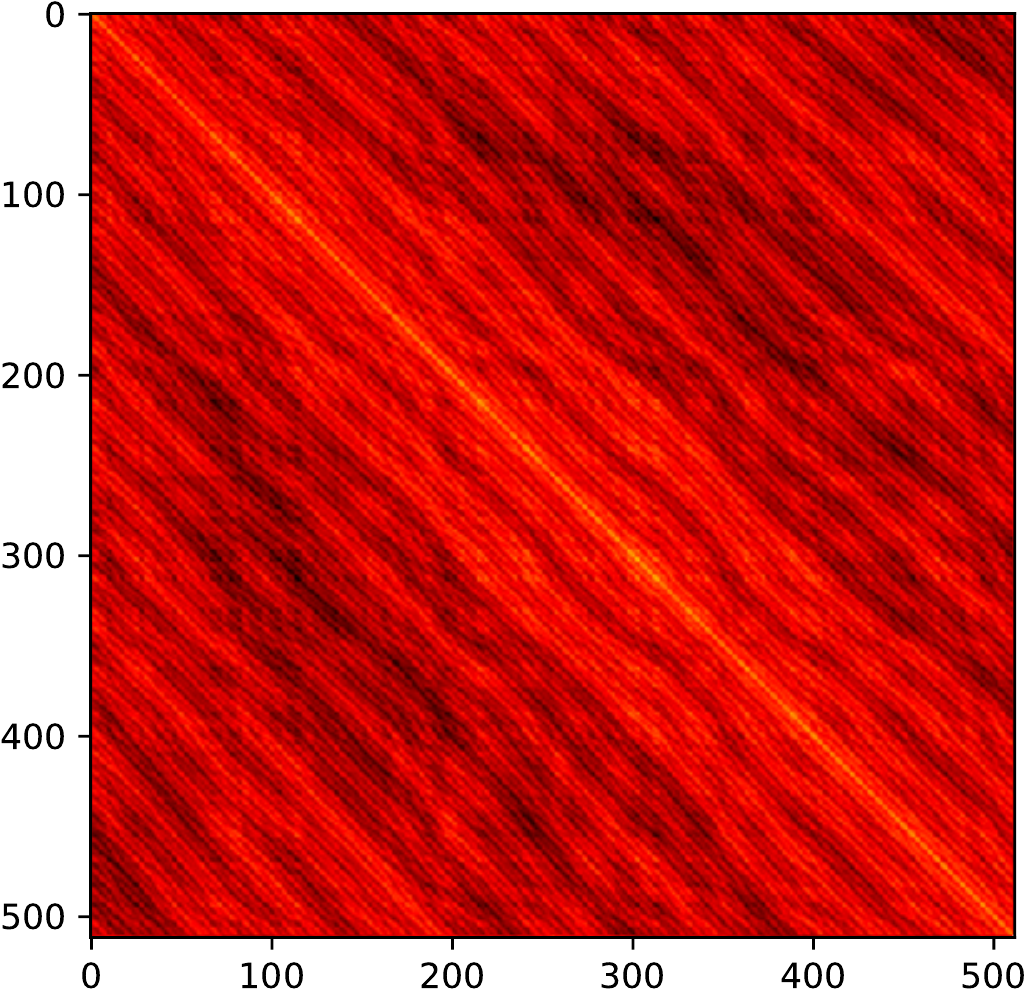}
	    \caption{ALBERT xlarge v1}
	\end{subfigure}
	\begin{subfigure}[t]{0.245\textwidth}
		\centering
		\includegraphics[width=\textwidth]{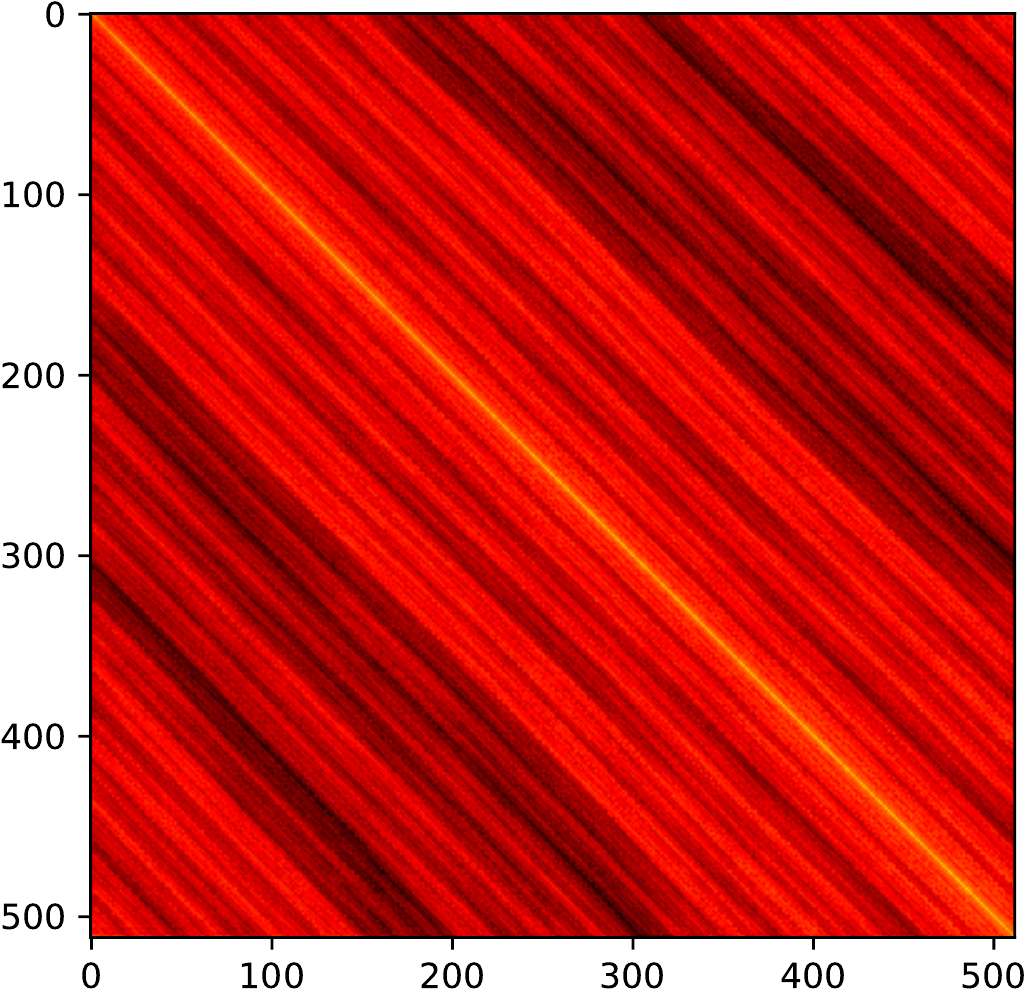}
	    \caption{ALBERT xxlarge v1}
	\end{subfigure}
	\begin{subfigure}[t]{0.245\textwidth}
		\centering
		\includegraphics[width=\textwidth]{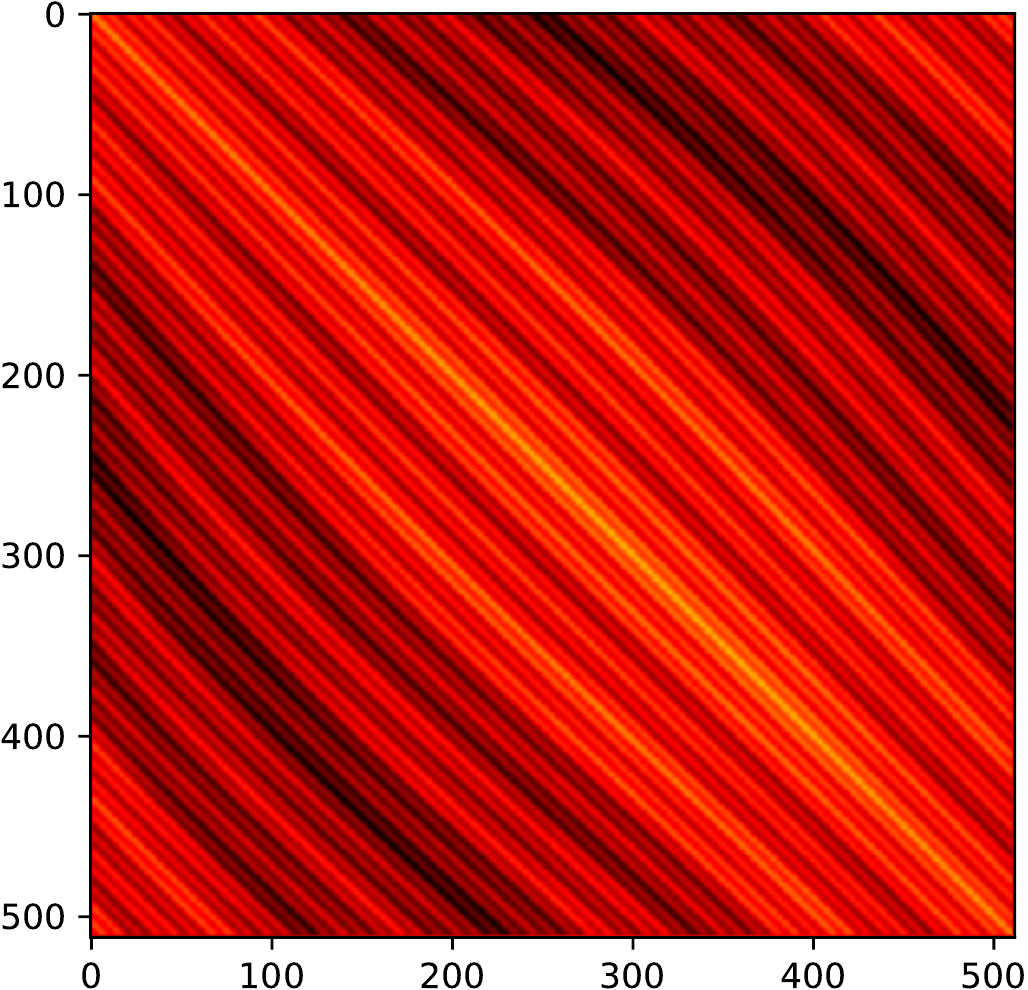}
	    \caption{ALBERT base v2}
	\end{subfigure}
	\begin{subfigure}[t]{0.245\textwidth}
		\centering
		\includegraphics[width=\textwidth]{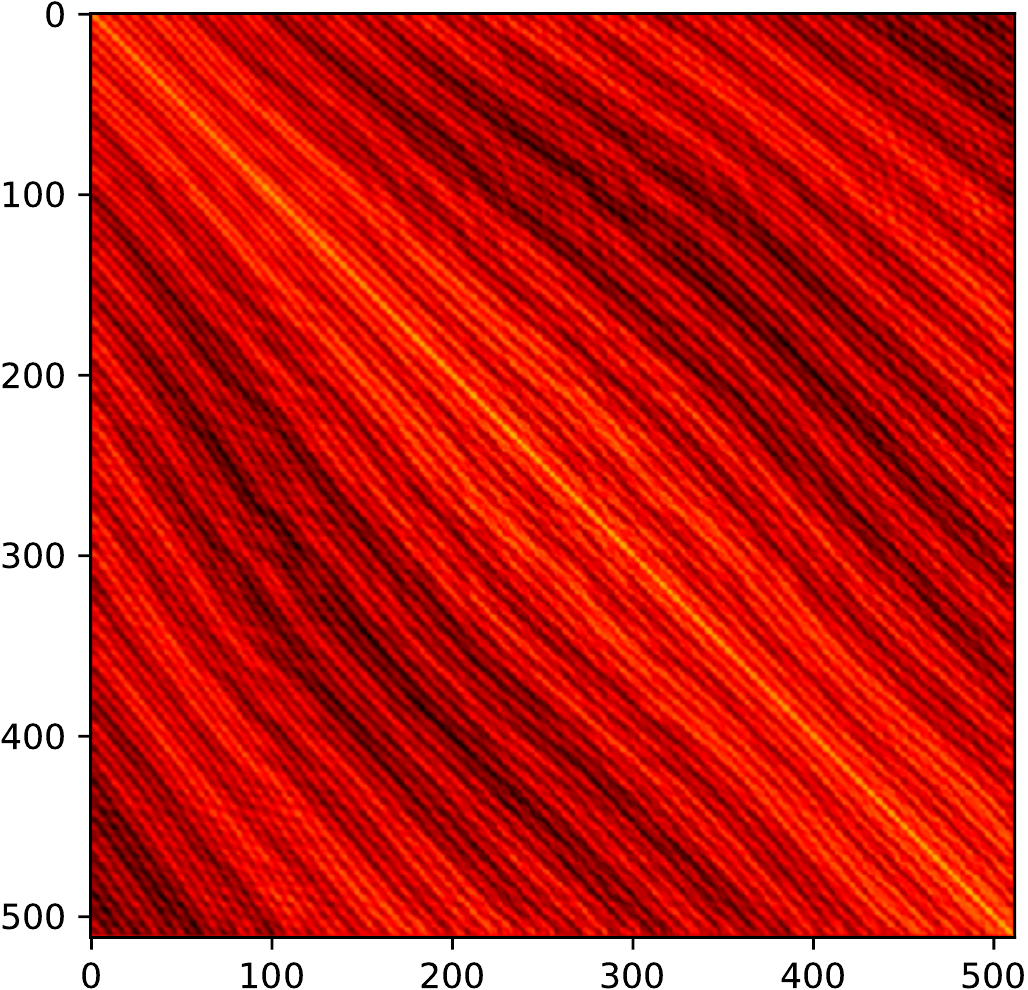}
	    \caption{ALBERT large v2}
	\end{subfigure}
	\begin{subfigure}[t]{0.245\textwidth}
		\centering
		\includegraphics[width=\textwidth]{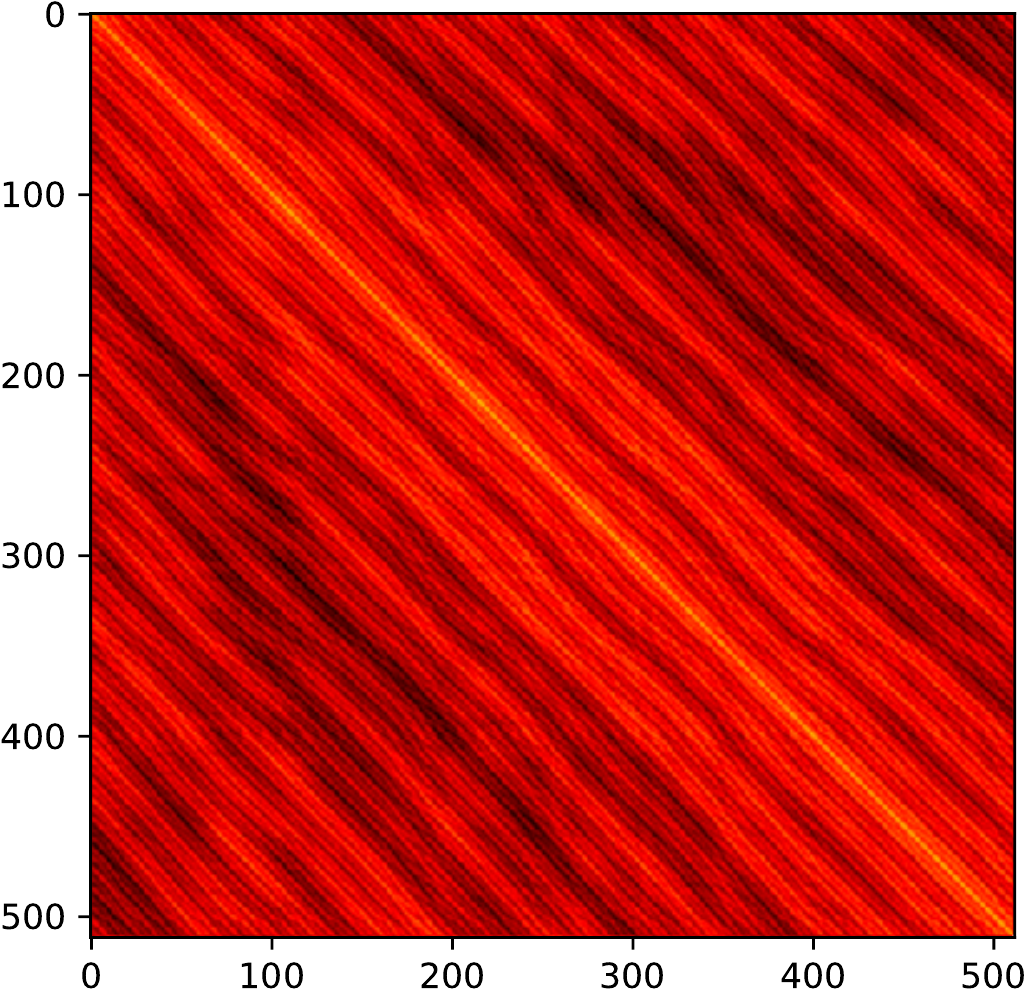}
	    \caption{ALBERT xlarge v2}
	\end{subfigure}
	\begin{subfigure}[t]{0.245\textwidth}
		\centering
		\includegraphics[width=\textwidth]{images/EpEpT/albert-xxlarge-v2-crop.pdf}
	    \caption{ALBERT xxlarge v2}
	\end{subfigure}
	\caption{Visualizations of the inner-product matrix $P = E_P E_P^T \in \real^{n \times n}$ for different ALBERT models \cite{lan2020albert}. We plot both v1 and v2 to show the progression towards increased Toeplitzness during training.}
	\label{fig:alberts}
    \vspace{-0.5em}
\end{figure*}

%% file: figures/multifix_suppl_merge_funcs.tex
\begin{figure*}[!t]
	\centering
	\hfill
	\begin{subfigure}[t]{0.485\textwidth}
		\centering
		\includegraphics[width=\textwidth, height=5.8cm]{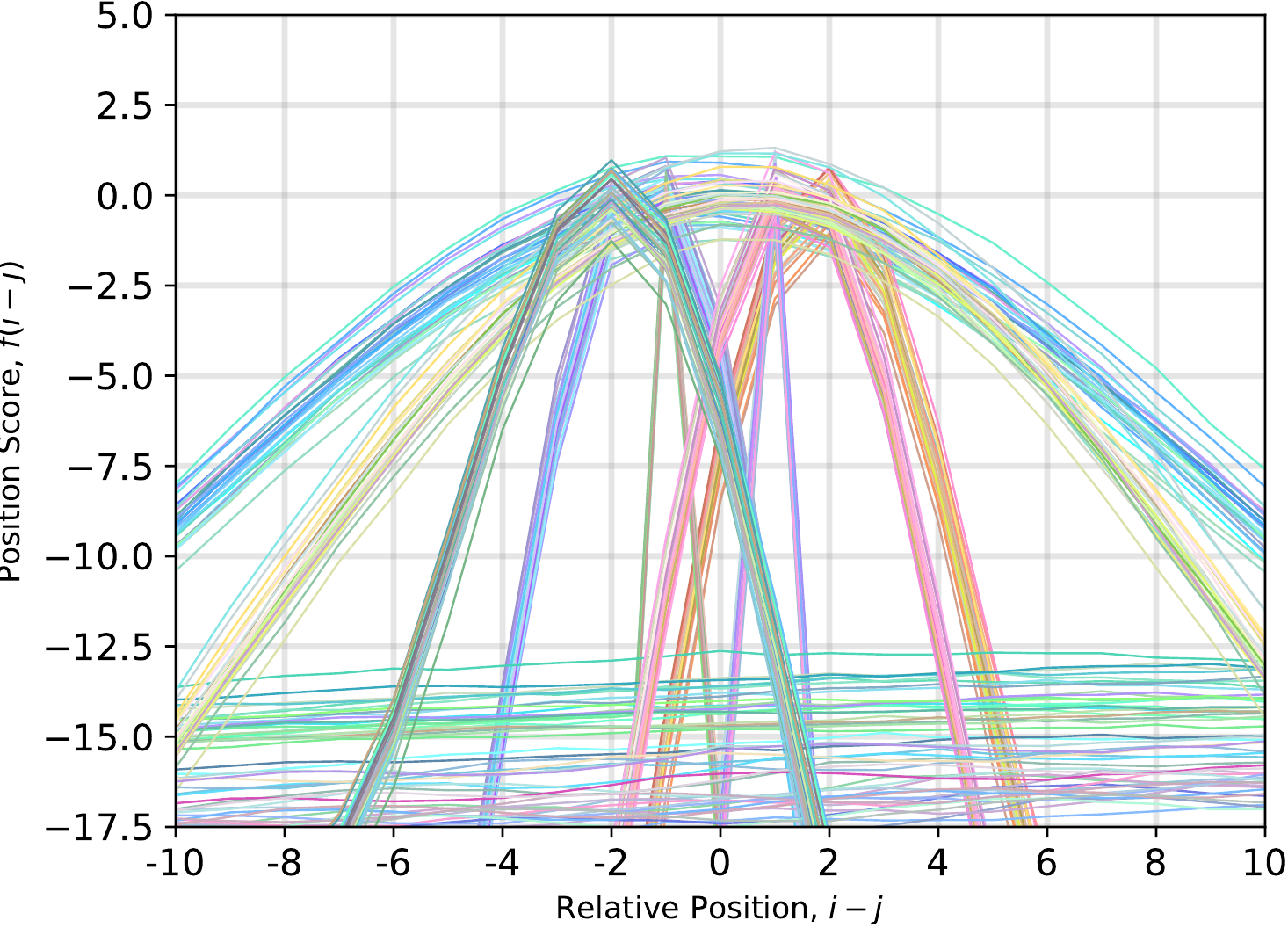}
	\end{subfigure}
	\hfill
	\hfill
	\begin{subfigure}[t]{0.485\textwidth}
		\centering
		\includegraphics[width=\textwidth, height=5.8cm]{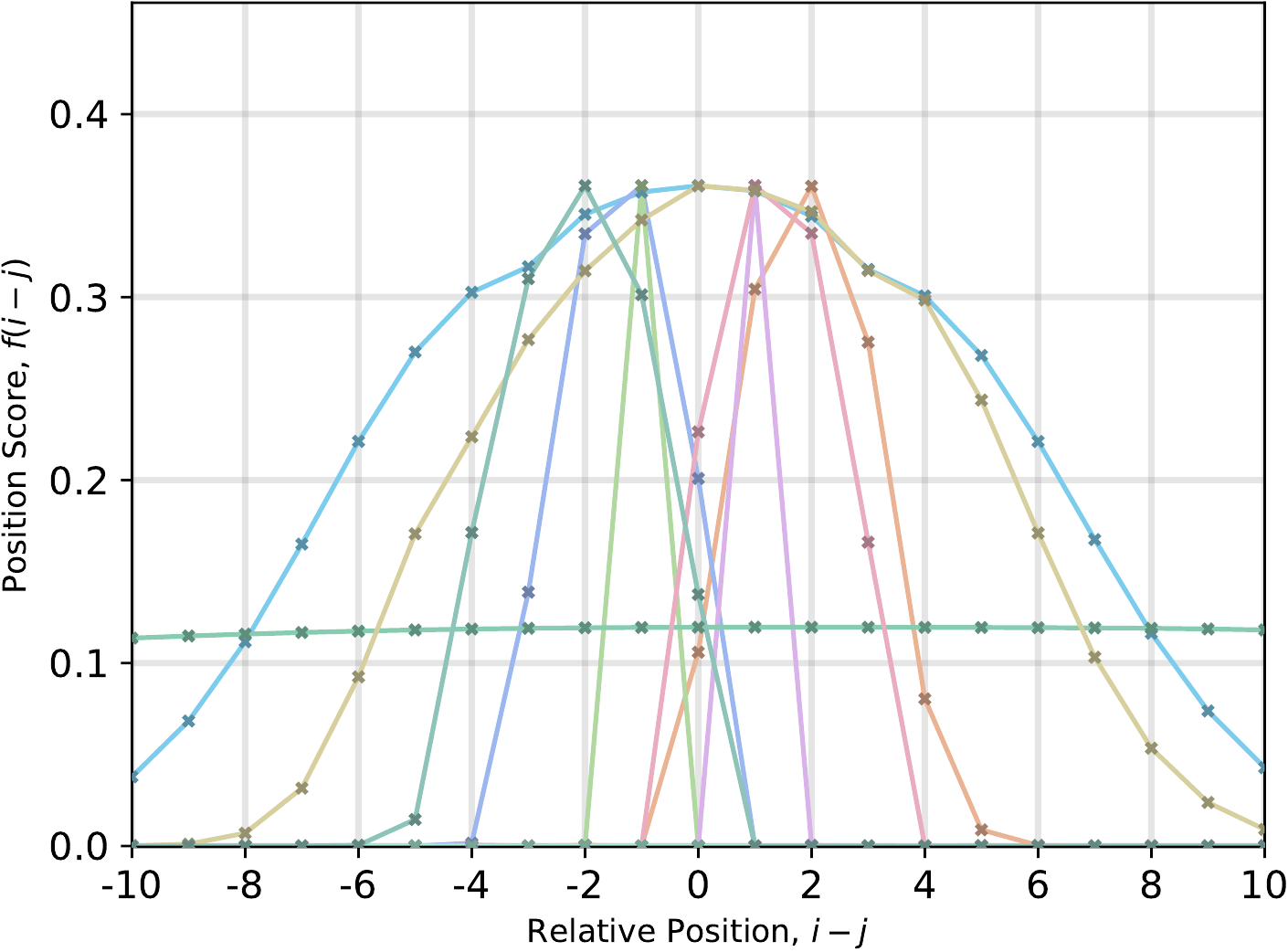}
	\end{subfigure}
	\hfill
	\caption{Positional responses of all attention heads. Sections through $\widehat{F}_P$ of ALBERT base v2, aligned to the main diagonal, (left) show similar profiles as the corresponding \ours{} scoring functions (right).
	Vertical axes differ due to 1) the scaling factor $\sqrt{d_k}$ and 2) softmax being invariant to vertical offset.
	}
	\label{fig:albertvstisa}
    \vspace{1.5em}
    \label{fig:ffunc_large}
	\centering
	\begin{subfigure}[t]{0.245\textwidth}
		\centering
		\includegraphics[width=\textwidth]{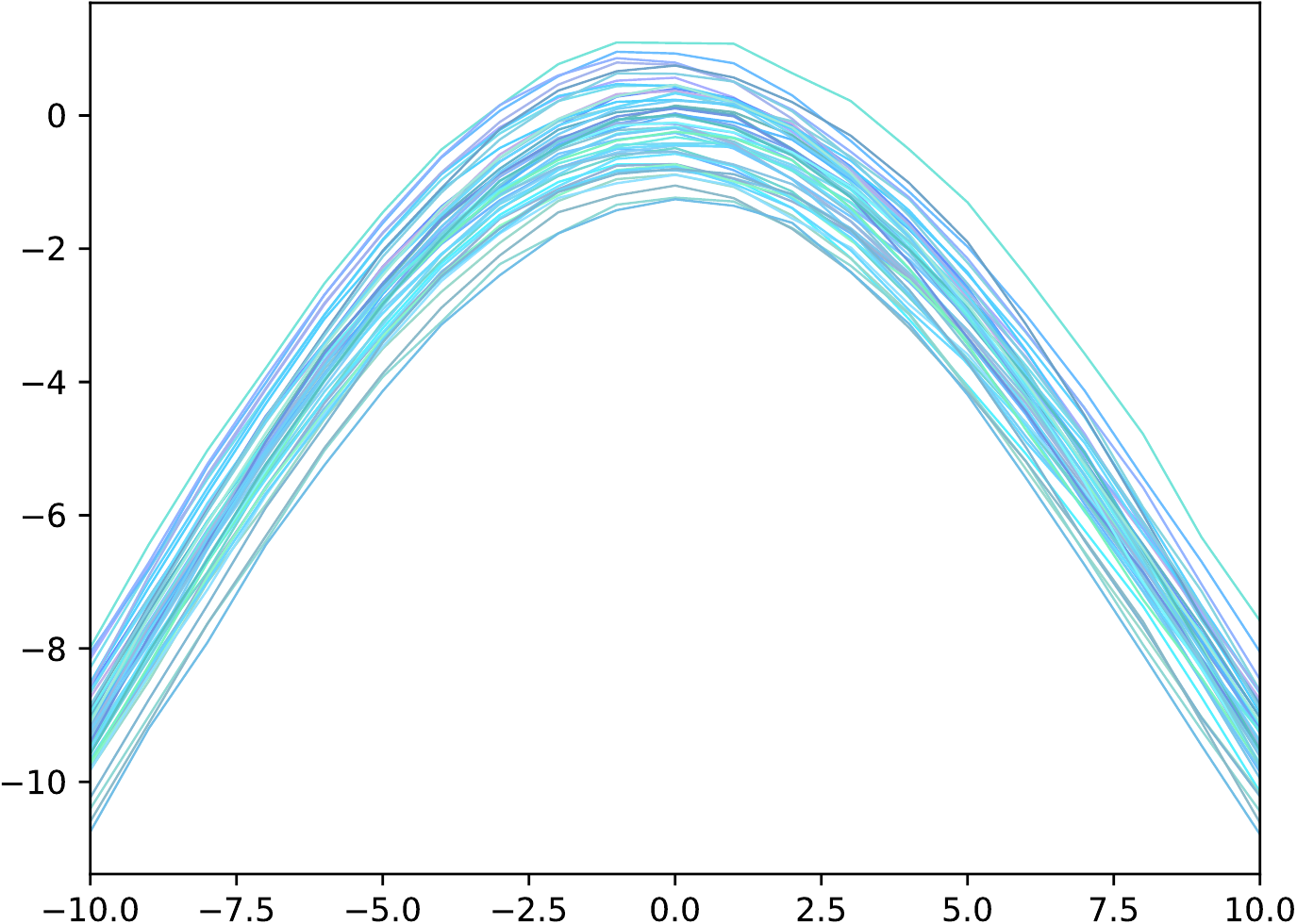}
		\caption{Attention head 1}
	\end{subfigure}
	\begin{subfigure}[t]{0.245\textwidth}
		\centering
		\includegraphics[width=\textwidth]{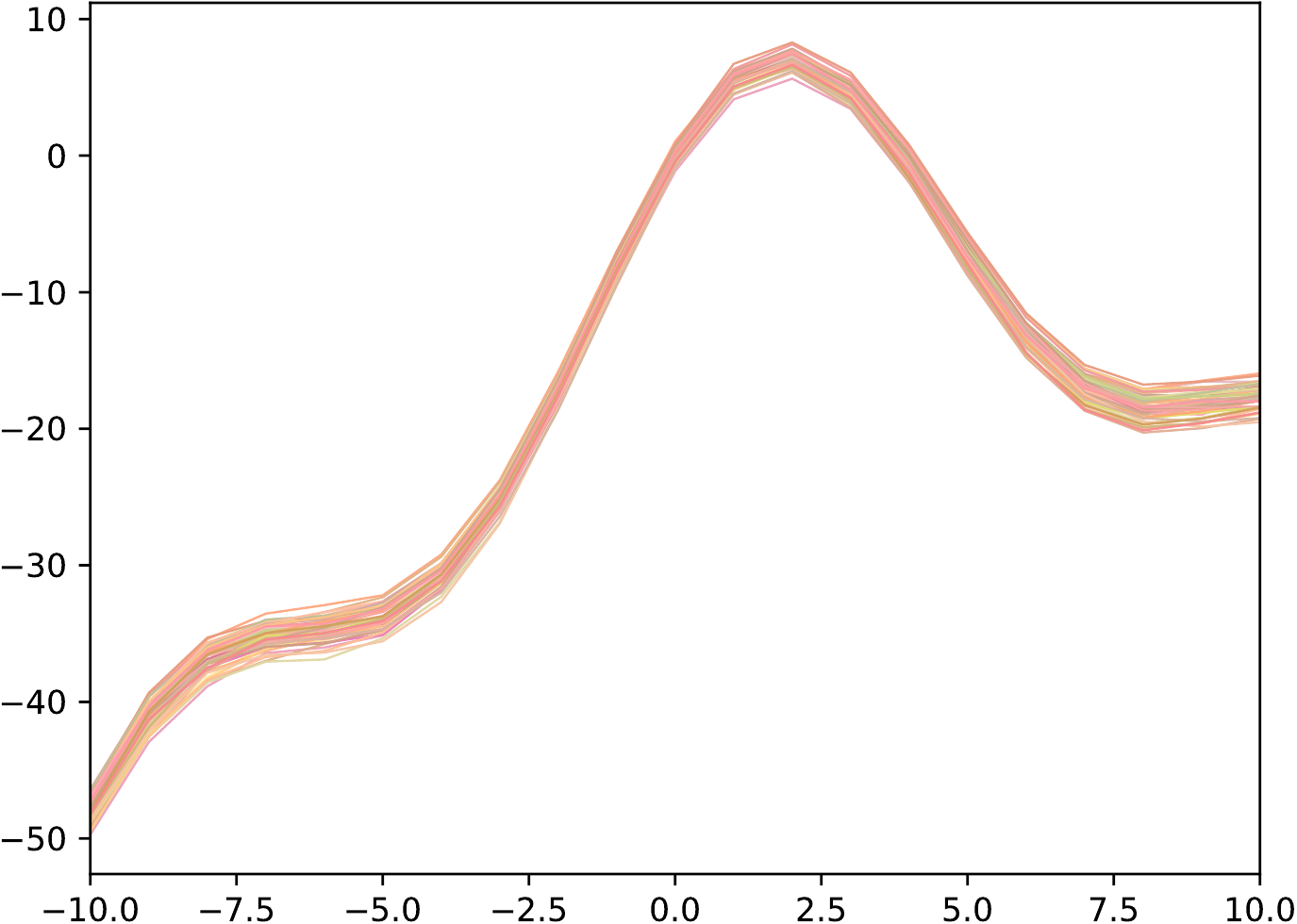}
		\caption{Attention head 2}
	\end{subfigure}
	\begin{subfigure}[t]{0.245\textwidth}
		\centering
		\includegraphics[width=\textwidth]{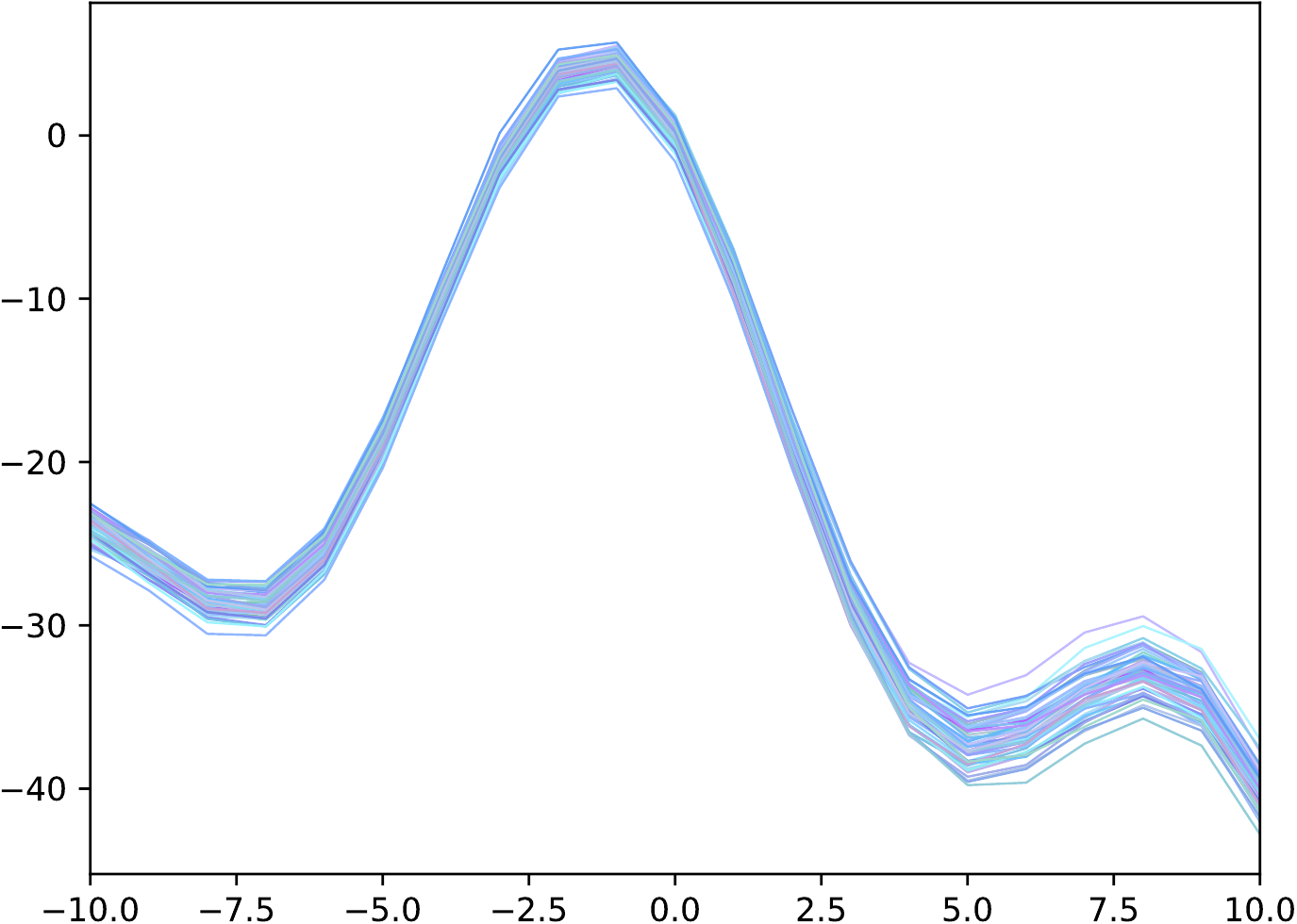}
		\caption{Attention head 3}
	\end{subfigure}
	\begin{subfigure}[t]{0.245\textwidth}
		\centering
		\includegraphics[width=\textwidth]{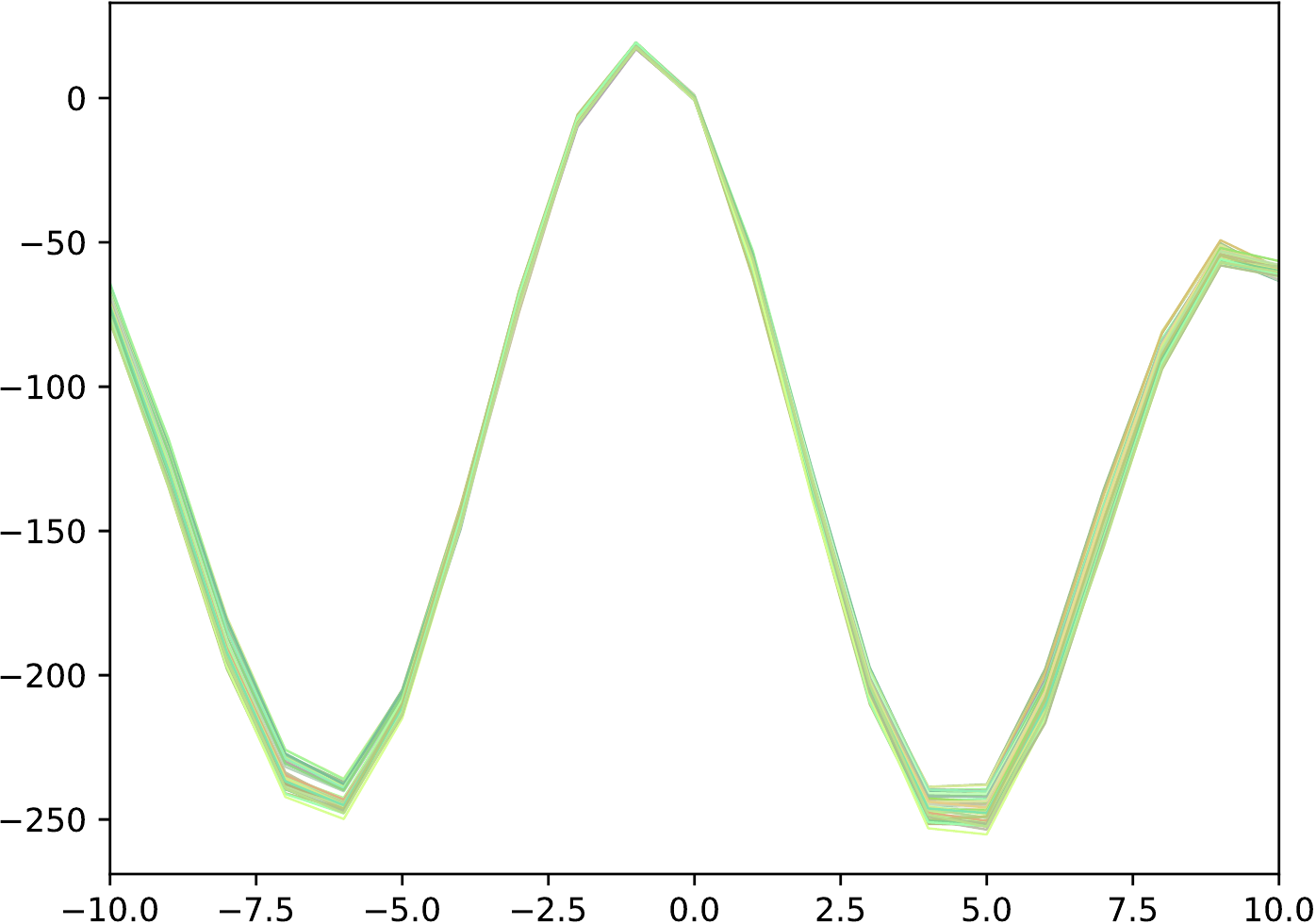}
		\caption{Attention head 4}
	\end{subfigure}
	\begin{subfigure}[t]{0.245\textwidth}
		\centering
		\includegraphics[width=\textwidth]{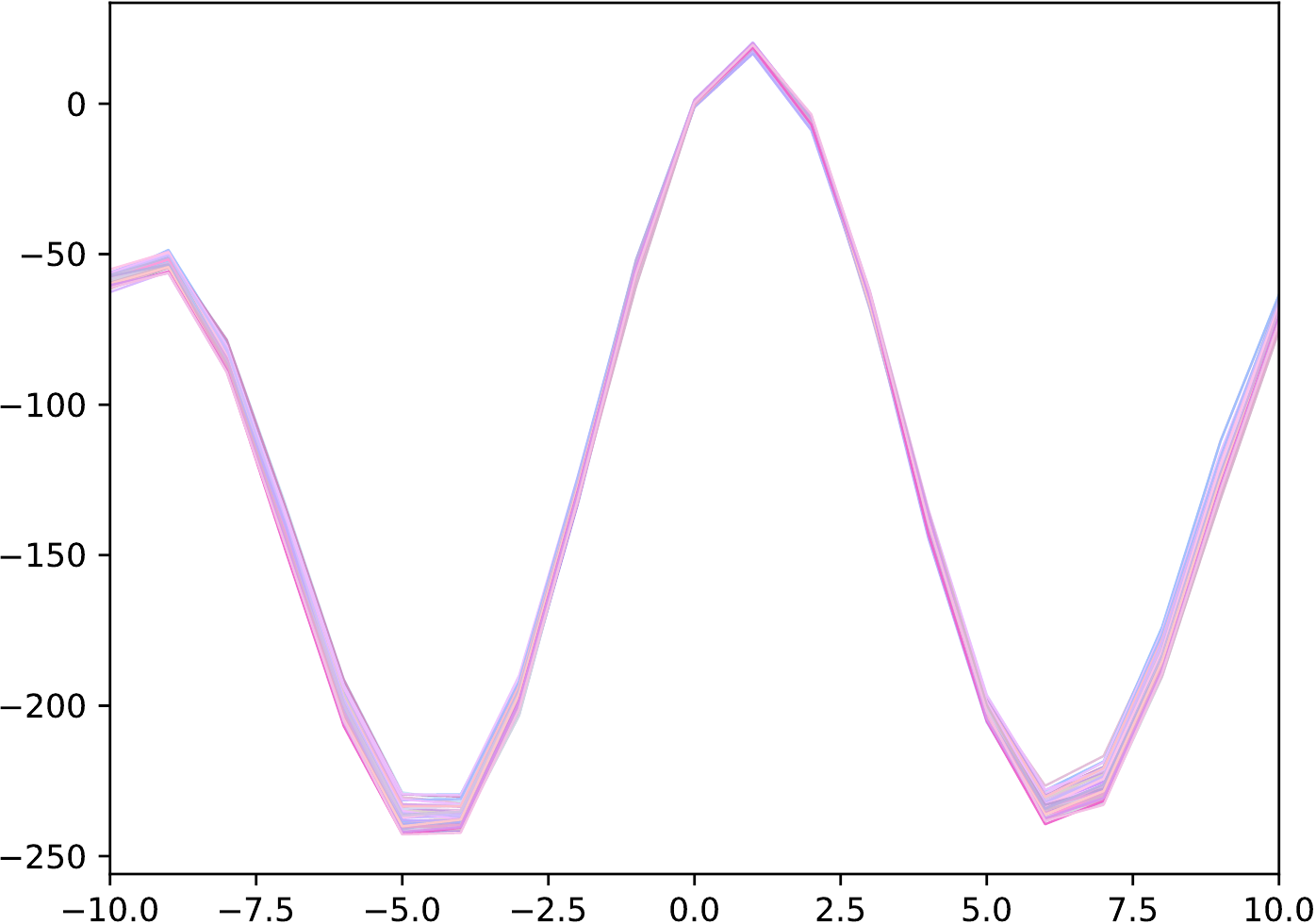}
		\caption{Attention head 5}
	\end{subfigure}
	\begin{subfigure}[t]{0.245\textwidth}
		\centering
		\includegraphics[width=\textwidth]{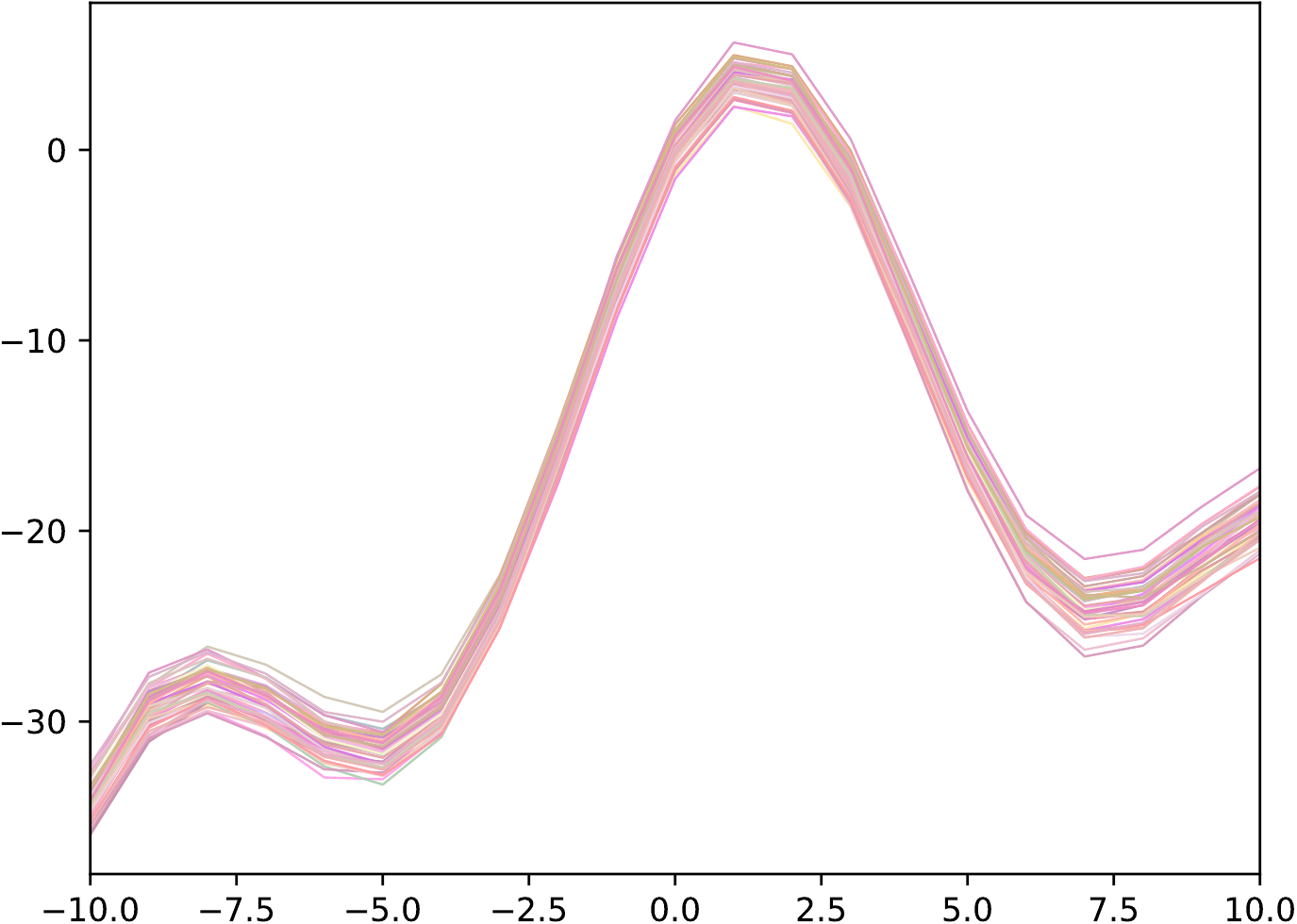}
		\caption{Attention head 6}
	\end{subfigure}
	\begin{subfigure}[t]{0.245\textwidth}
		\centering
		\includegraphics[width=\textwidth]{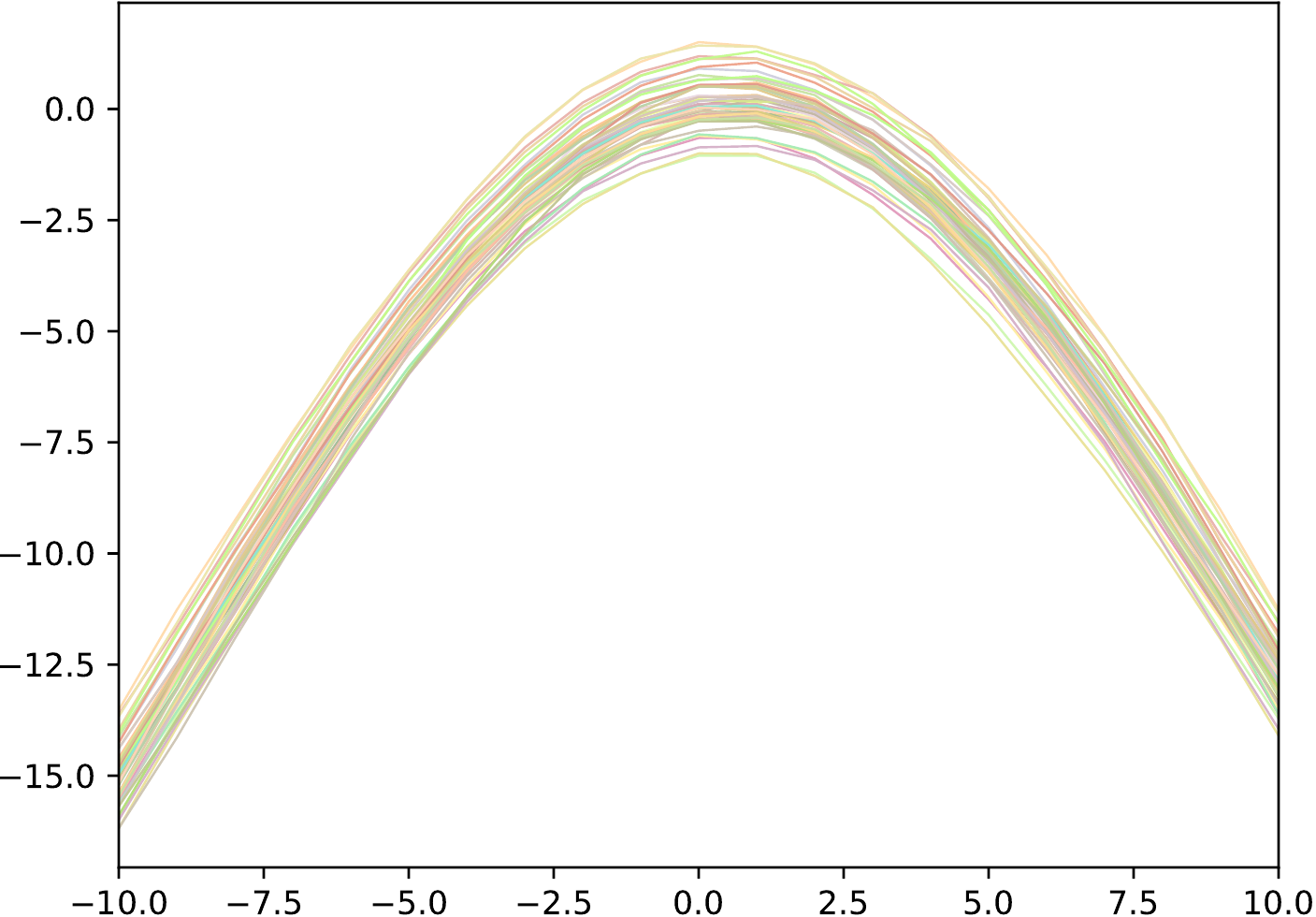}
		\caption{Attention head 7}
	\end{subfigure}
	\begin{subfigure}[t]{0.245\textwidth}
		\centering
		\includegraphics[width=\textwidth]{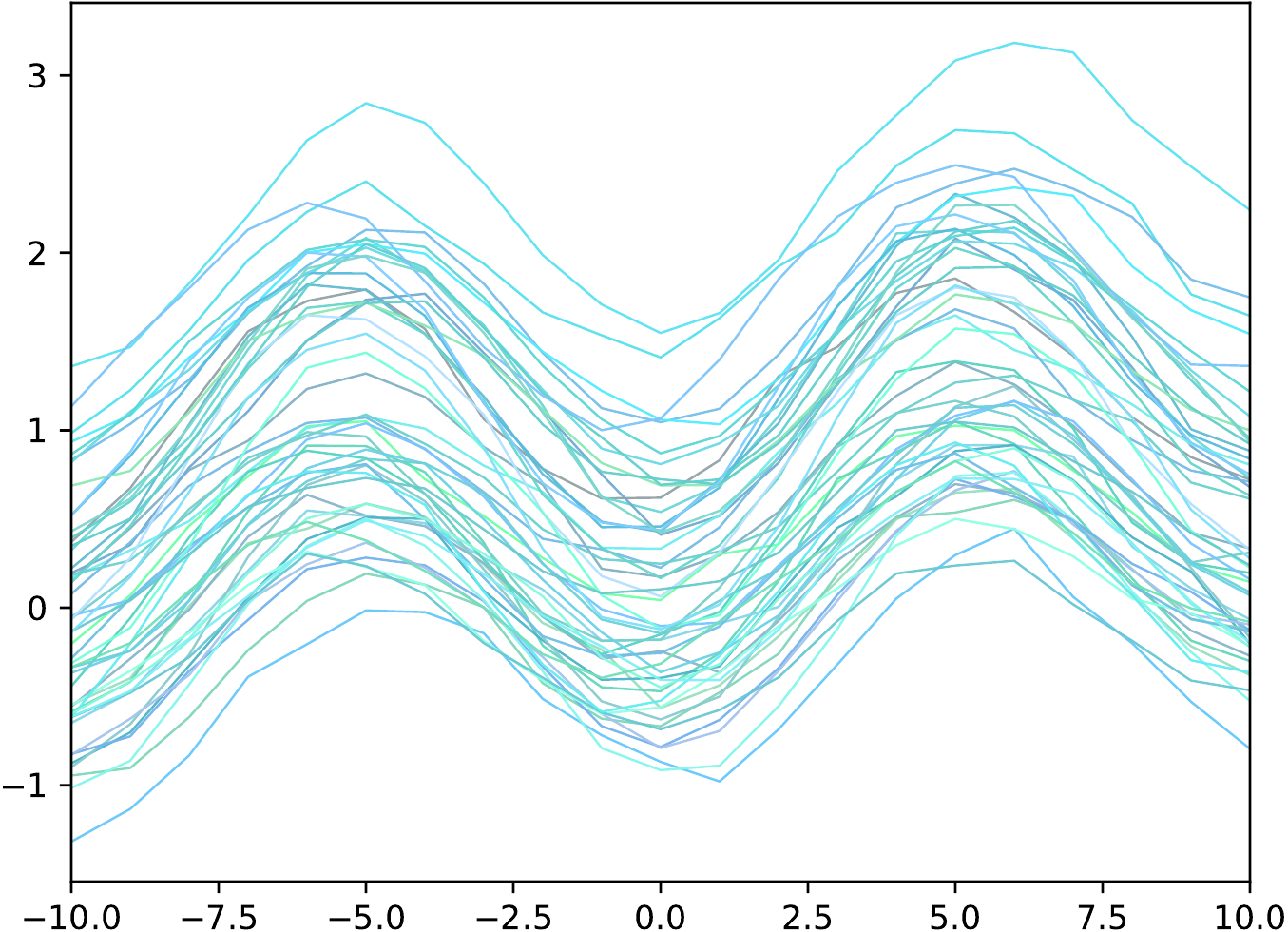}
		\caption{Attention head 8}
	\end{subfigure}
	\begin{subfigure}[t]{0.245\textwidth}
		\centering
		\includegraphics[width=\textwidth]{images/ALBERT_Attention_Function/attention_weight_functions_08-crop.pdf}
		\caption{Attention head 9}
	\end{subfigure}
	\begin{subfigure}[t]{0.245\textwidth}
		\centering
		\includegraphics[width=\textwidth]{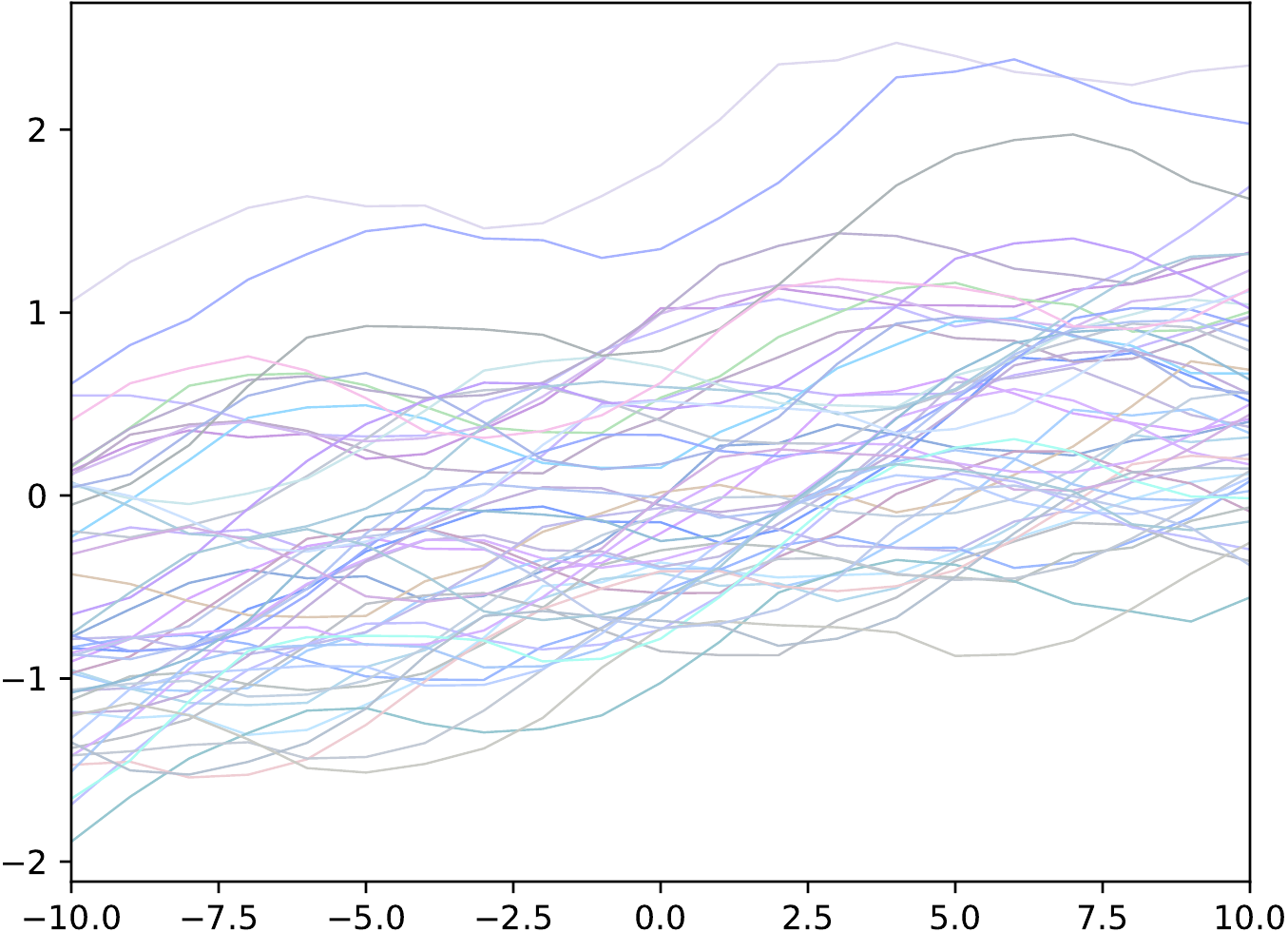}
		\caption{Attention head 10}
	\end{subfigure}
	\begin{subfigure}[t]{0.245\textwidth}
		\centering
		\includegraphics[width=\textwidth]{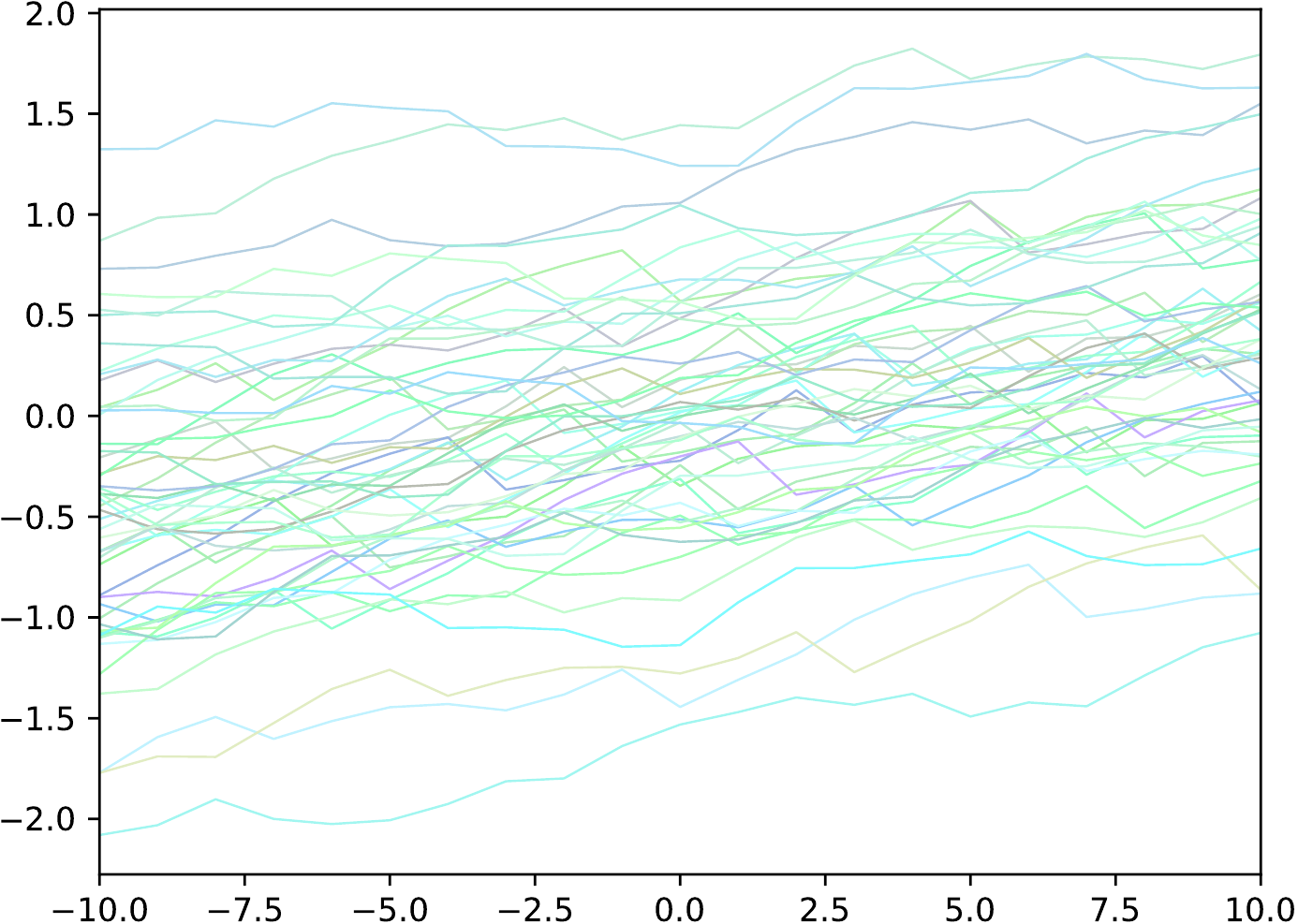}
		\caption{Attention head 11}
	\end{subfigure}
	\begin{subfigure}[t]{0.245\textwidth}
		\centering
		\includegraphics[width=\textwidth]{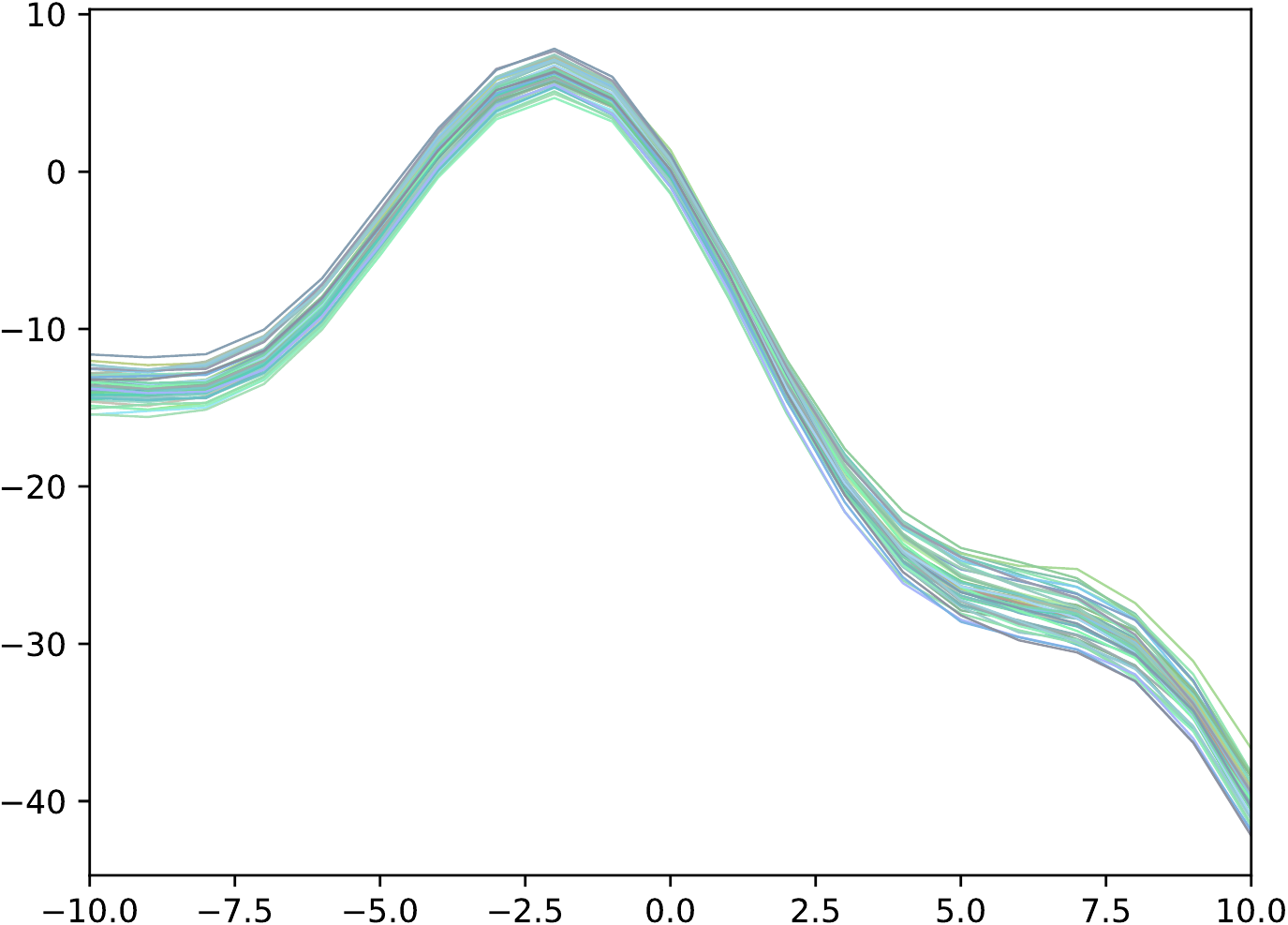}
		\caption{Attention head 12}
	\end{subfigure}
	\caption{Rows from the positional attention matrices $\widehat{F}_P$ for all ALBERT base v2 attention heads, centered on the main diagonal. Note that the vertical scale generally differs between plots. The plots are essentially aligned sections through the matrices in Fig.\ \ref{fig:allheadmatrices}, but zoomed in to show details over short relative distances since this is where the main peak(s) are located, and the highest values are by far the most influential on softmax attention.}
	\label{fig:allheadrows}
\end{figure*}

%% file: figures/multifix_suppl5_albert_regular_att_matrix.tex
\begin{figure*}[ht!]
	\centering
	\begin{subfigure}[t]{0.245\textwidth}
		\centering
		\includegraphics[width=\textwidth]{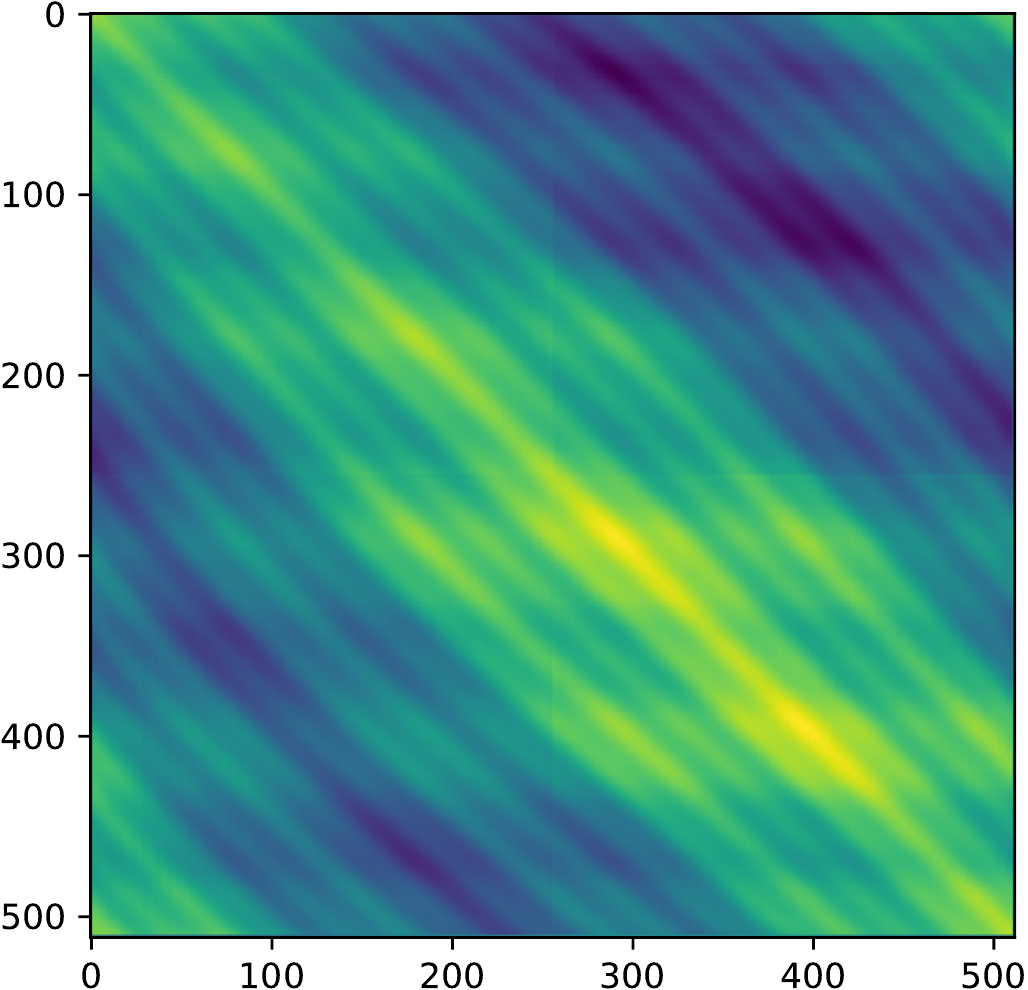}
		\caption{Attention head 1}
	\end{subfigure}
	\begin{subfigure}[t]{0.245\textwidth}
		\centering
		\includegraphics[width=\textwidth]{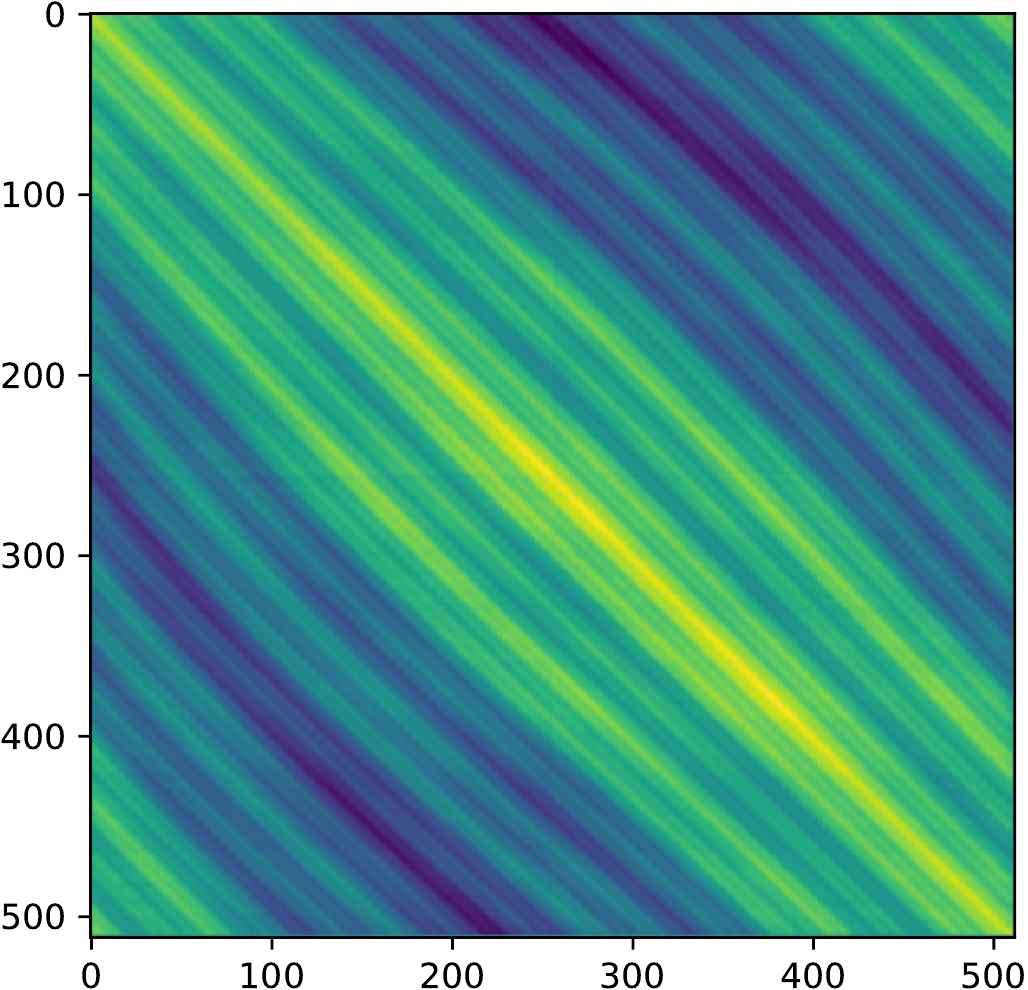}
		\caption{Attention head 2}
	\end{subfigure}
	\begin{subfigure}[t]{0.245\textwidth}
		\centering
		\includegraphics[width=\textwidth]{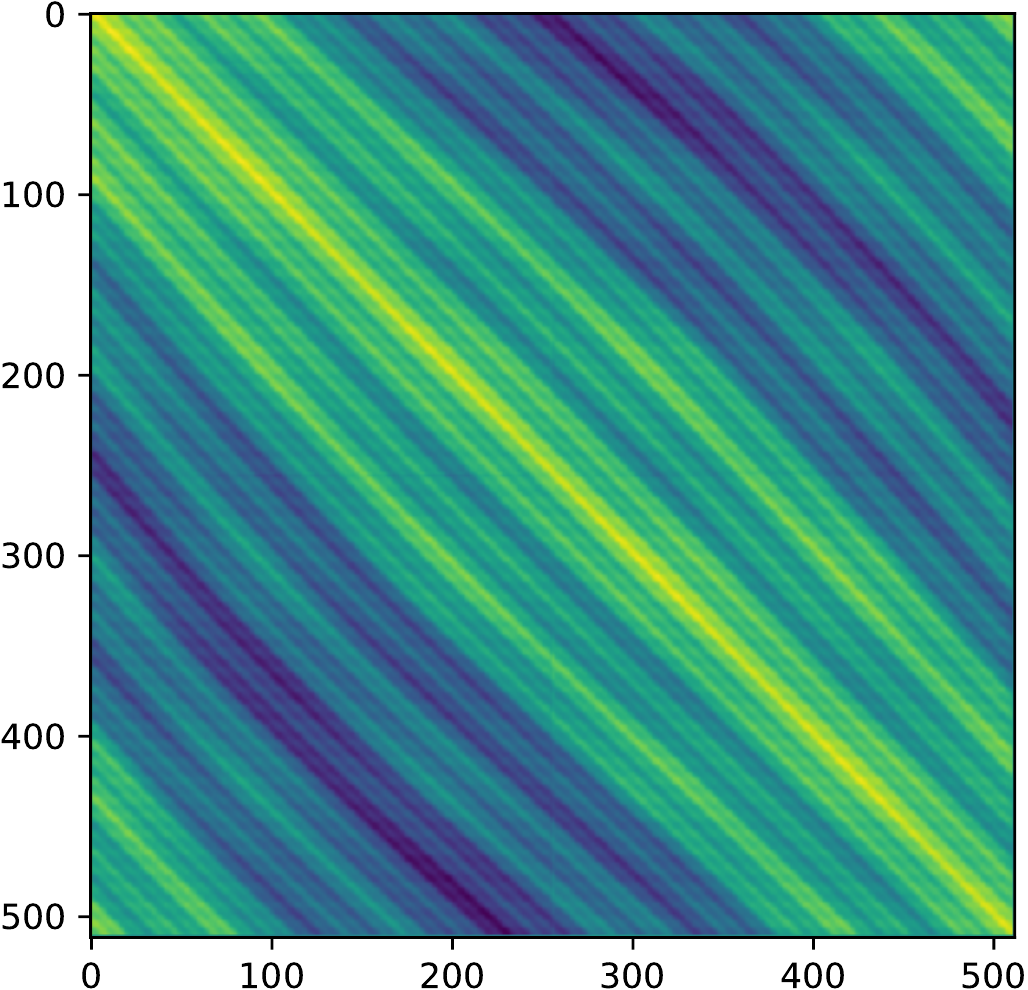}
		\caption{Attention head 3}
	\end{subfigure}
	\begin{subfigure}[t]{0.245\textwidth}
		\centering
		\includegraphics[width=\textwidth]{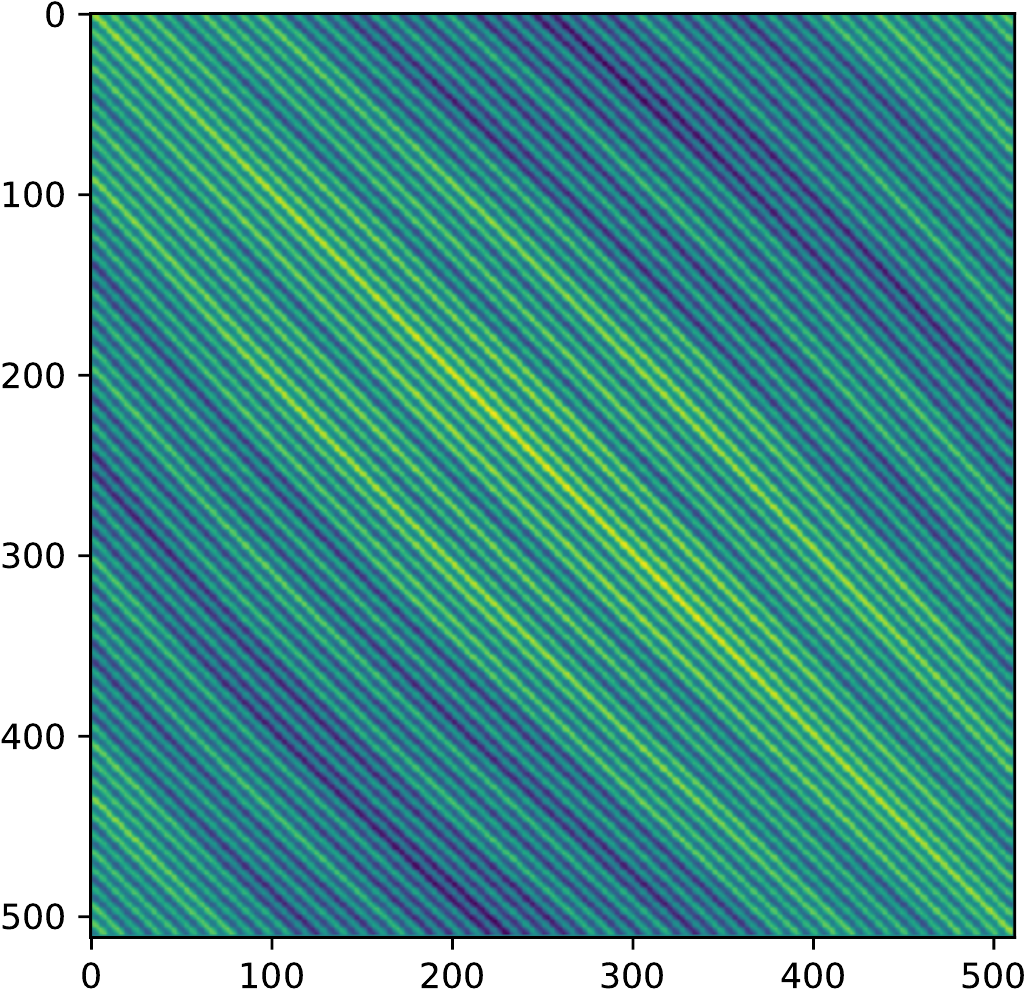}
		\caption{Attention head 4}
	\end{subfigure}
	\begin{subfigure}[t]{0.245\textwidth}
		\centering
		\includegraphics[width=\textwidth]{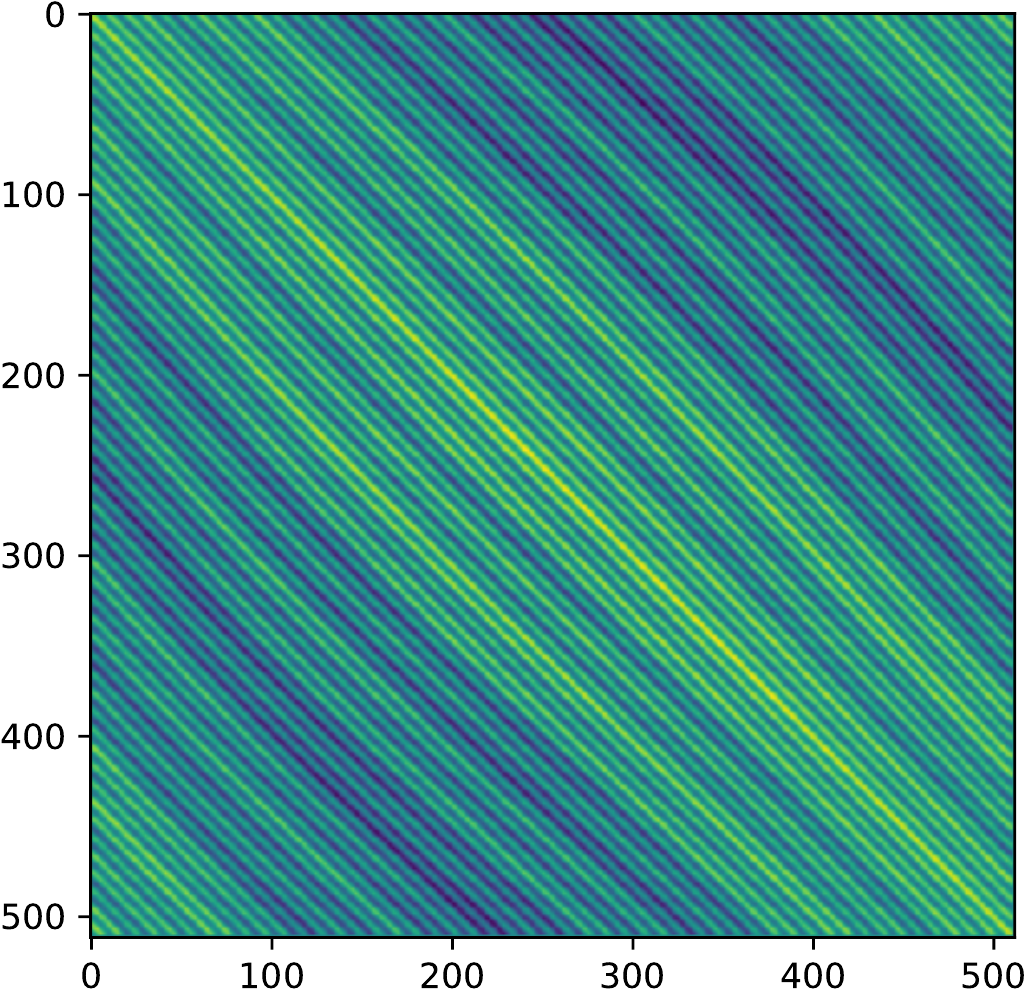}
		\caption{Attention head 5}
	\end{subfigure}
	\begin{subfigure}[t]{0.245\textwidth}
		\centering
		\includegraphics[width=\textwidth]{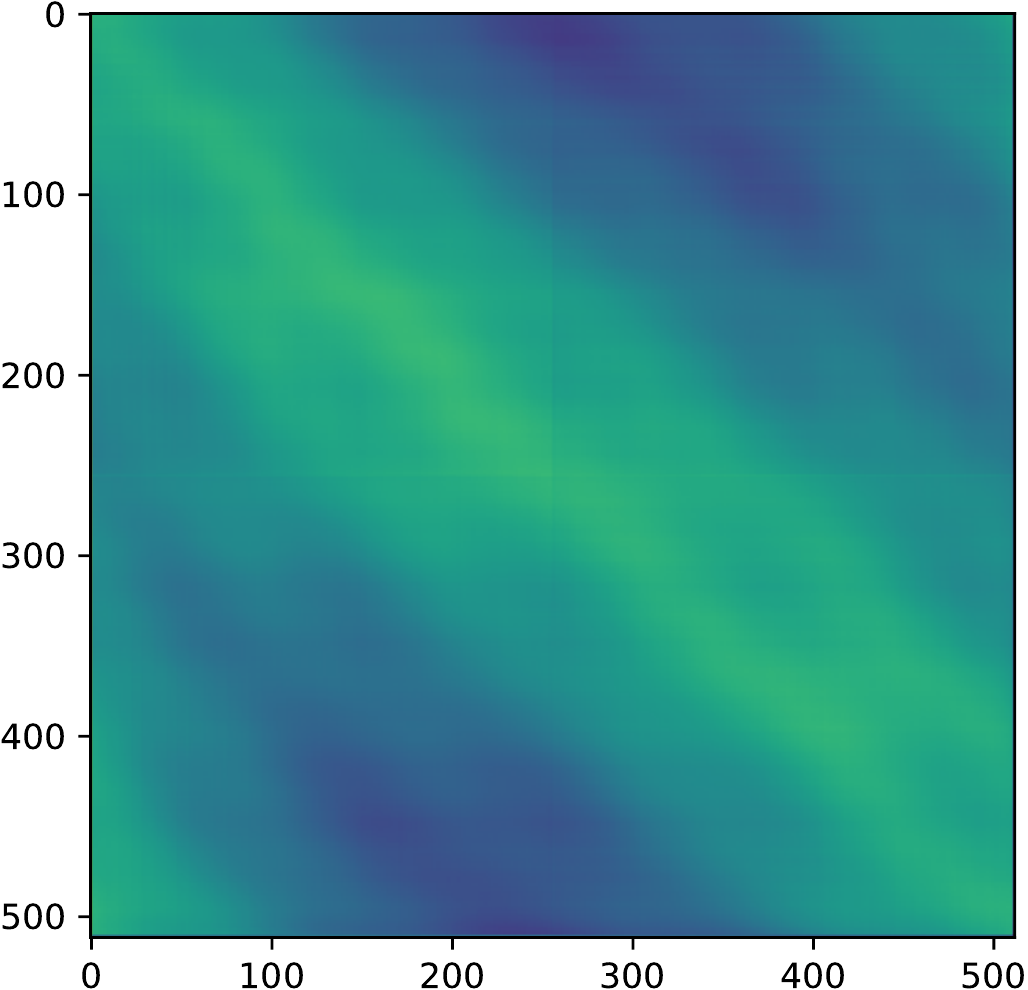}
		\caption{Attention head 6}
	\end{subfigure}
	\begin{subfigure}[t]{0.245\textwidth}
		\centering
		\includegraphics[width=\textwidth]{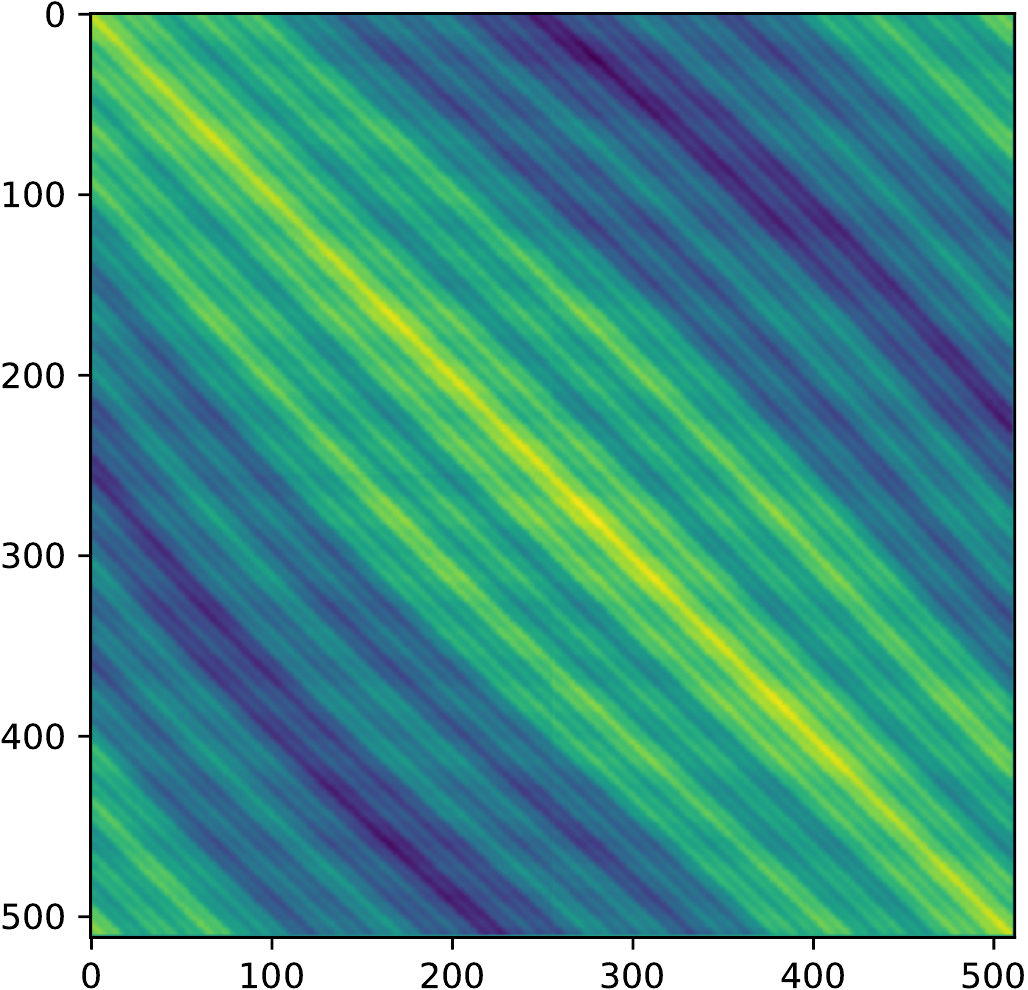}
		\caption{Attention head 7}
	\end{subfigure}
	\begin{subfigure}[t]{0.245\textwidth}
		\centering
		\includegraphics[width=\textwidth]{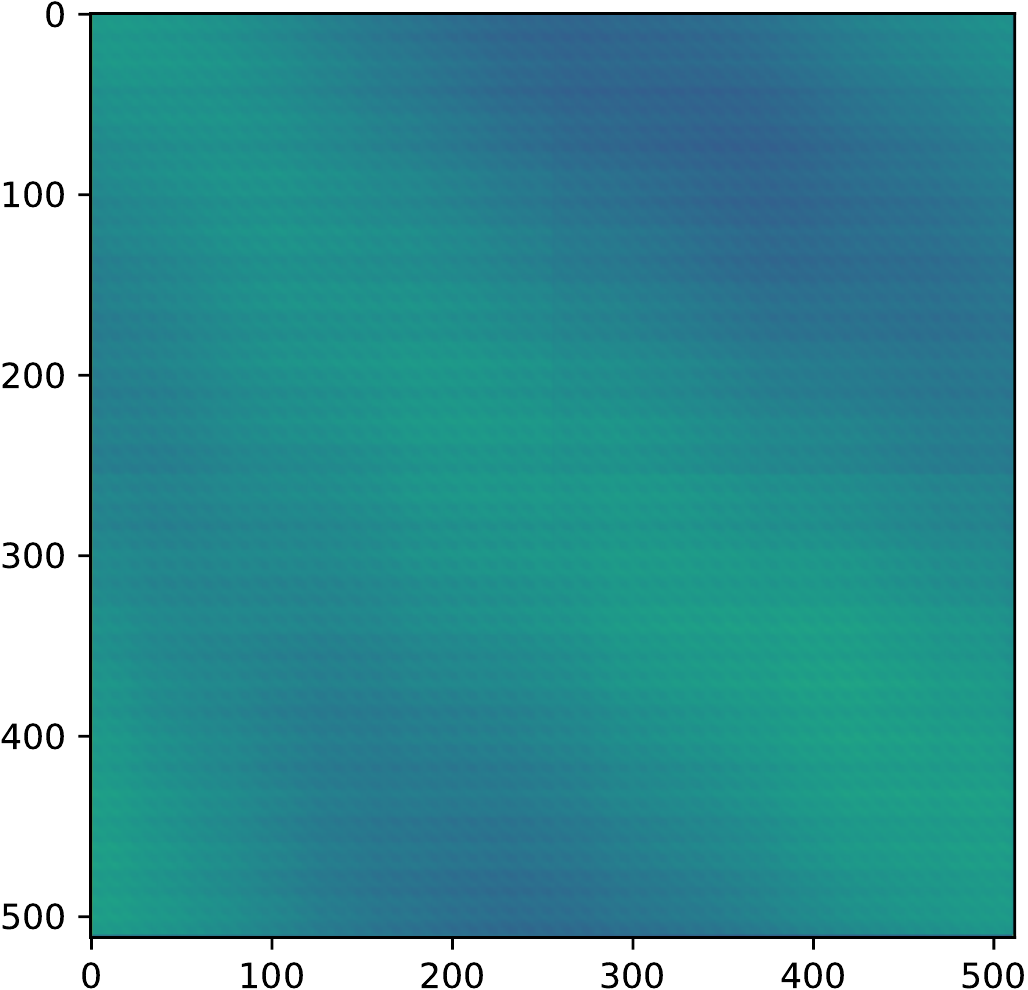}
		\caption{Attention head 8}
	\end{subfigure}
	\begin{subfigure}[t]{0.245\textwidth}
		\centering
		\includegraphics[width=\textwidth]{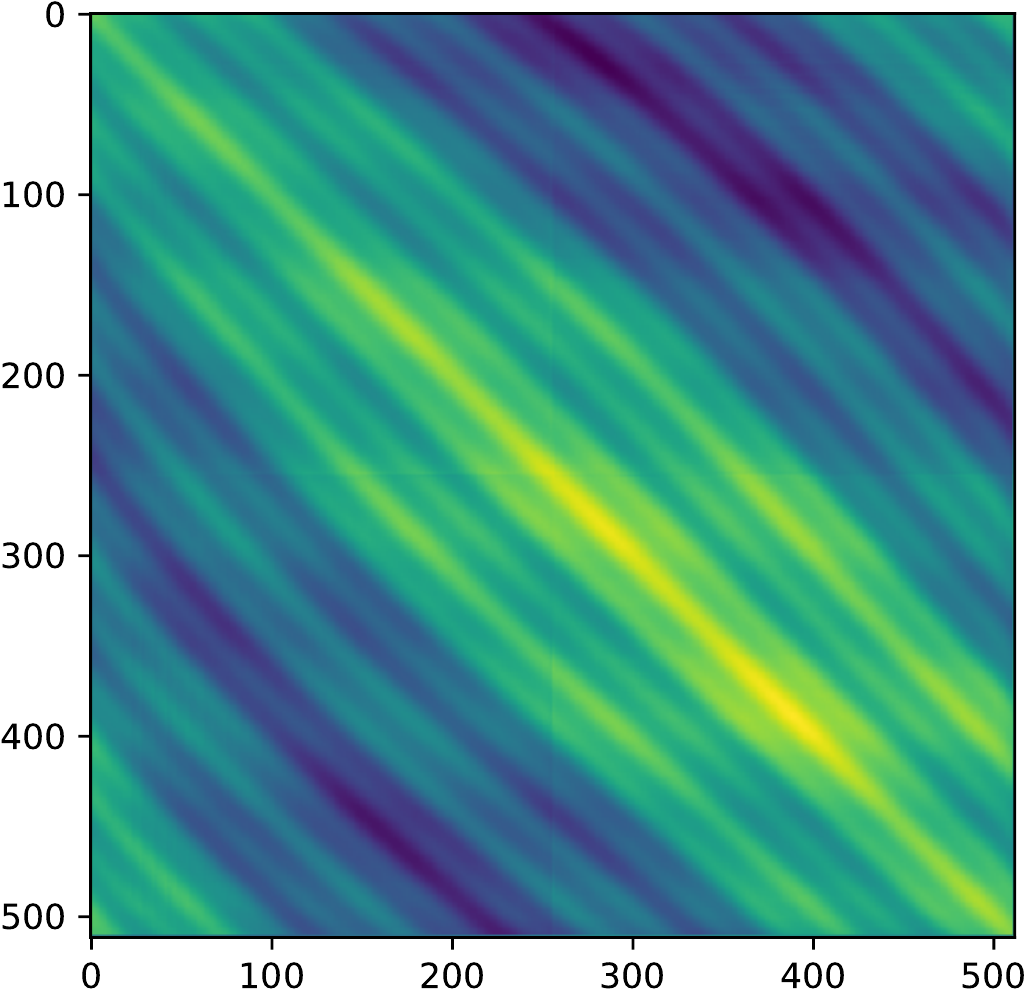}
		\caption{Attention head 9}
	\end{subfigure}
	\begin{subfigure}[t]{0.245\textwidth}
		\centering
		\includegraphics[width=\textwidth]{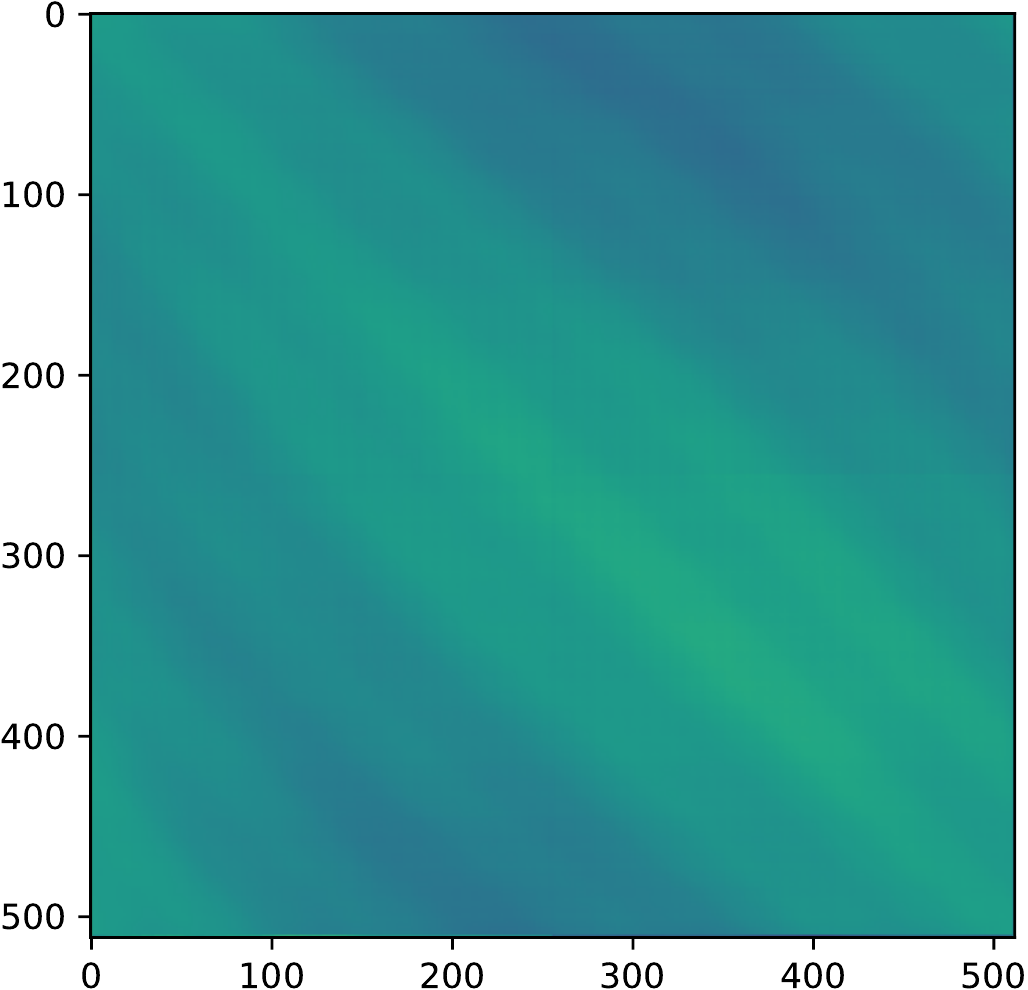}
		\caption{Attention head 10}
	\end{subfigure}
	\begin{subfigure}[t]{0.245\textwidth}
		\centering
		\includegraphics[width=\textwidth]{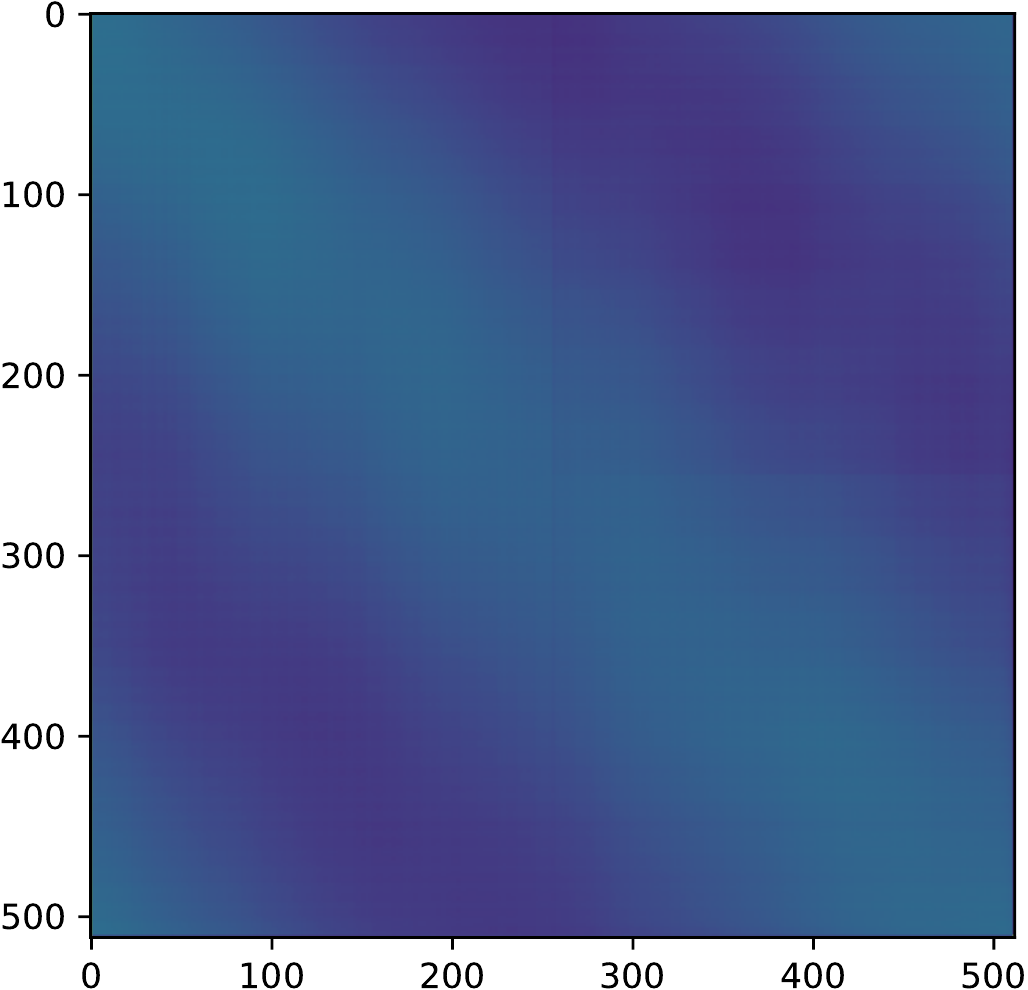}
		\caption{Attention head 11}
	\end{subfigure}
	\begin{subfigure}[t]{0.245\textwidth}
		\centering
		\includegraphics[width=\textwidth]{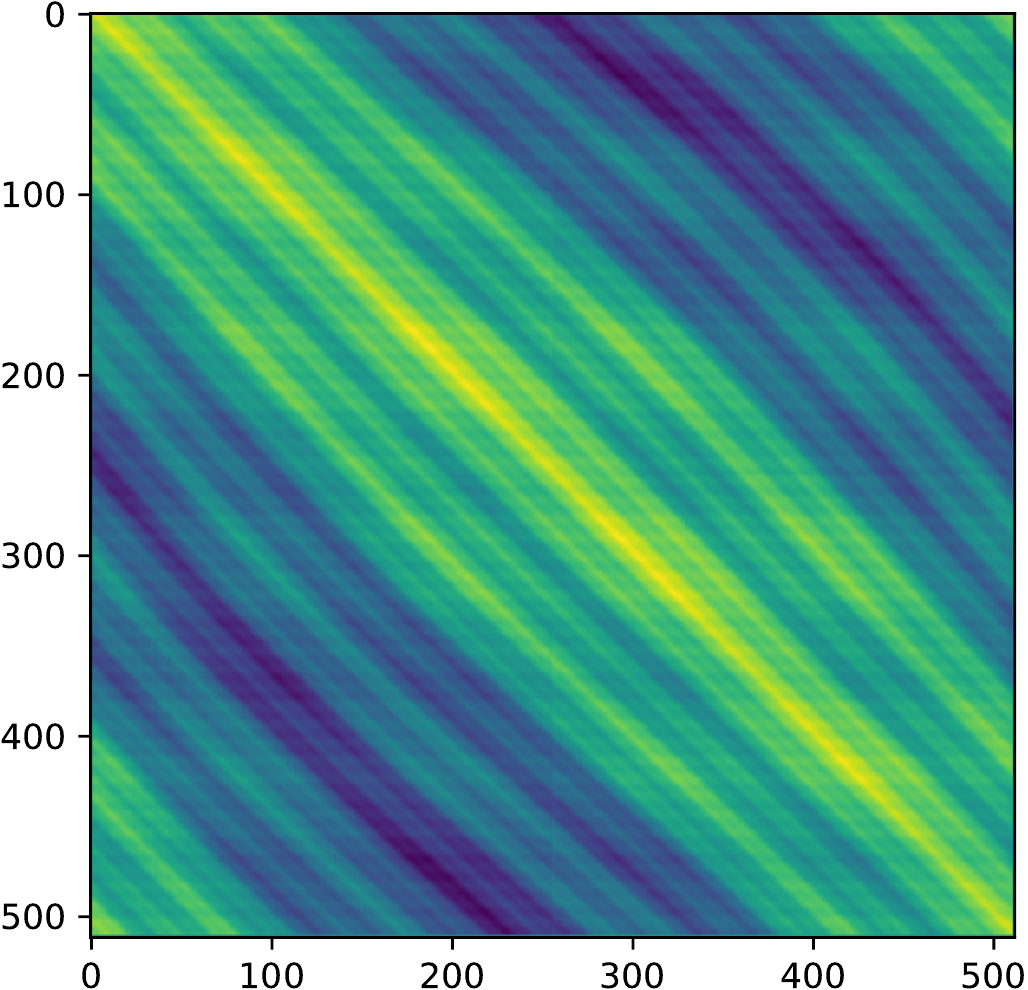}
		\caption{Attention head 12}
	\end{subfigure}
	\caption{Values extracted from the positional attention matrices for all ALBERT base v2 first-layer attention heads. Some heads are seen to be sensitive to position, while others are not. Note that these visualizations deliberately use a different color scheme from other (red) matrices, to emphasize the fact that the matrices visualized here represent a different phenomenon and are not inner products.}
	\label{fig:allheadmatrices}
    \vspace{-0.5em}
\end{figure*}